%% file: bare_conf_compsoc.tex
\documentclass[conference,compsoc]{IEEEtran}
\usepackage{threeparttable}
\usepackage{tikz}
\usepackage{epsfig}
\usepackage{graphicx}
\usepackage{url} 
\usepackage{makecell}

\usepackage{textcomp}

\usepackage[linesnumbered, ruled]{algorithm2e}
\usepackage[ruled,linesnumbered]{algorithm2e}
\usepackage{xurl}
\PassOptionsToPackage{table,dvipsnames}{xcolor}
\usepackage{bbding}
\usepackage{booktabs}       
\usepackage{amsfonts}       
\usepackage{nicefrac}       
\usepackage{soul}
\usepackage{makecell}
\usepackage{hyperref}
\usepackage{microtype}      
\usepackage{multirow}
\usepackage{booktabs}
\usepackage{adjustbox}
\usepackage{epstopdf}
\usepackage{enumitem}
\usepackage{mathrsfs} 
\usepackage{bm} 
\usepackage{tabularx}
\usepackage[most]{tcolorbox}
\usepackage{color}
\usepackage{xspace}
\usepackage{hyperref}
\usepackage{colortbl}
\usepackage{stmaryrd}
\usepackage{fvextra}
\usepackage{epsfig}
\usepackage{graphicx}
\usepackage{multirow}
\usepackage{adjustbox}
\usepackage{array}
\usepackage{enumitem}
\usepackage{multirow}
\usepackage{makecell}
\usepackage{hyperref}
\usepackage{ragged2e}
\usepackage{threeparttable}
\usepackage{relsize}
\usepackage{xspace}
\usepackage{algpseudocode}
\usepackage[normalem]{ulem}
\usepackage{bbm}
\usepackage[T1]{fontenc}
\usepackage[utf8]{inputenc}
\usepackage{tcolorbox}
\usepackage{url} 
\usepackage{algpseudocode}
\usepackage{colortbl}
\usepackage{cellspace}
\usepackage{pifont}
\usepackage{tabularx}

\usepackage[most]{tcolorbox}

\definecolor{summaryblue}{RGB}{42,78,150}
\definecolor{obsgreen}{RGB}{34,135,76}
\definecolor{summarybg}{RGB}{247,248,251}

\newtcolorbox{summarybox}[1]{
  enhanced,
  breakable,
  colback=summarybg,
  colframe=summarybg,
  boxrule=0pt,
  borderline west={2pt}{0pt}{#1},
  left=5pt,
  right=5pt,
  top=3pt,
  bottom=3pt,
  boxsep=0pt,
  before skip=5pt,
  after skip=5pt
}

\newcommand{\soksummary}[2]{%
\begin{summarybox}{summaryblue}
\noindent\textbf{Summary #1:}~\emph{#2}
\end{summarybox}
}

\definecolor{secRisk}{HTML}{FADBD8} 
\definecolor{secMit}{HTML}{D9E2F3}
\newcommand{\halfsquare}{%
\begin{tikzpicture}[scale=0.18,baseline={(0,0)}]
  \fill (0,0) rectangle (0.5,1);
  \draw (0,0) rectangle (1,1);
\end{tikzpicture}%
}

\newcommand{\emptysquare}{%
\begin{tikzpicture}[scale=0.18,baseline={(0,0)}]
  \draw (0,0) rectangle (1,1);
\end{tikzpicture}%
}

\newcommand{\fullsquare}{%
\begin{tikzpicture}[scale=0.18,baseline={(0,0)}]
  \fill (0,0) rectangle (1,1);
  \draw (0,0) rectangle (1,1); 
\end{tikzpicture}%
}

\definecolor{lightred}{RGB}{254, 244, 236}
\definecolor{lightgreen}{RGB}{241, 248, 236}
\definecolor{lightblue}{RGB}{236, 243, 250}

\DefineVerbatimEnvironment{promptcode}{Verbatim}{
  breaklines=true,      
  breakanywhere=true,   
  breaksymbol={},
  fontsize=\small,
}

\ifCLASSOPTIONcompsoc
  \usepackage[nocompress]{cite}
\else
  \usepackage{cite}
\fi
\ifCLASSINFOpdf
\else
\fi

\usepackage{eso-pic}

\begin{document}

%
\title{SoK: Security and Privacy of Foundation-Model-Powered Robots}

\author{
\IEEEauthorblockN{Xueluan Gong$^{1}$, Chen Chen$^{1*}$, Jinxin Liu$^1$, Qian Wang$^2$, and Kwok-Yan Lam$^1$}

$^1$College of Computing and Data Science, Nanyang Technological University, Singapore\\
$^2$School of Cyber Science and Engineering, Wuhan University, China\\
$^*$Corresponding author \\
\{xueluan.gong, chen.chen, kwokyan.lam\}@ntu.edu.sg,
jinxin001@e.ntu.edu.sg, \\\{qianwang\}@whu.edu.cn
}


%


\maketitle









\begin{abstract}
Foundation models are reshaping robotics by enabling robots to interpret open-ended instructions, reason over multimodal contexts, and operate in complex, open-world environments. However, their integration also introduces security and privacy (S\&P) risks that extend beyond the FMs themselves to embodied execution pipelines, supporting ecosystems, and broader governance impacts. 
Existing literature reviews provide valuable insights but often focus on specific FM types, risk categories, mitigation strategies, or trust boundaries. Consequently, the field lacks a unified structure for analyzing where risks originate, how they propagate across robotic systems, and where mitigations should intervene.
To address this gap, we propose a progressive F-E-S-G structural boundary framework for analyzing the S\&P of FM-powered robots. The framework comprises four layers: the \emph{Foundation model layer} (F), \emph{Embodied system layer} (E), \emph{Supporting ecosystem layer} (S), and \emph{Governance impact layer} (G). Building on this structure, we develop a multi-level taxonomy that organizes prior studies along three levels: F-E-S-G trust boundary, security-privacy concerns, and risk-mitigation perspectives. We further annotate each study using fine-grained coding attributes, including
target, lifecycle stage, mechanism, system access, and effect.
Guided by this framework and taxonomy, we systematize 96 papers. Our analysis uncovers multiple threat patterns, defense mismatches, and evaluation gaps that
are difficult to identify from a single-boundary perspective.
Based on these findings, we identify open challenges and future directions to provide a research agenda for developing secure, privacy-preserving, and responsibly governed FM-powered robotic systems.



\end{abstract}

\IEEEpeerreviewmaketitle

\section{Introduction}
Robotics is emerging as an increasingly important technological frontier, with the potential to transform manufacturing, public service, and a broad range of physical-world applications.
To realize this potential, traditional robotics systems are typically developed through carefully engineered pipelines for perception, planning, and control. Despite achieving strong performance, their capabilities are generally limited to well-defined tasks, controlled environments, and narrow operating assumptions \cite{bai2025towards}. Recently, this paradigm has been reshaped by the emergence of foundation models (FMs) \cite{khan2025foundation}. For instance, Large Language Models (LLMs) are increasingly integrated to support instruction understanding and high-level reasoning \cite{wang2025large}; vision-language models (VLMs) facilitate multimodal grounding and scene interpretation \cite{pantazopoulos2025towards,miao2026survey}; and vision-language-action models (VLAs) further enable robots to directly map visual signals to executable actions \cite{turgunbaev2025perception}. As a result, foundation-model-powered (FM-powered) robots are evolving into general-purpose embodied agents capable of interpreting open-ended instructions, reasoning over complex contexts, and operating in diverse physical environments. 

This transition also introduces substantial security and privacy (S\&P) concerns. Traditional robots are mainly exposed to cyber-physical and software-level threats, such as sensor spoofing and insecure communication channels \cite{zhu2024earbench}. FM-powered robots inherit these risks while introducing FM-induced vulnerabilities, including hallucinated reasoning, goal misinterpretation, poisoned contexts, or compromised model components
\cite{lu2024poex,wang2025trojanrobot,robey2025jailbreaking}. These threats may propagate across the embodied pipeline, potentially leading to physical harm or privacy violations. 
For example, Robey et al. \cite{robey2025jailbreaking} demonstrate that LLM-controlled robots, including the Clearpath Jackal UGV and Unitree Go2, can be jailbroken into executing harmful behaviors. Zhang et al. \cite{zhang2024badrobot} further show that embodied LLM agents can be manipulated through voice-based interactions to perform unsafe actions. Without effective safeguards, such vulnerabilities may result in serious real-world consequences. Therefore, S\&P must be treated as core requirements throughout the development and deployment of FM-powered robotic systems \cite{chen2024trustworthy}.

\input{Table/survey_com}

Recently, a growing body of work has examined S\&P issues in FM-powered robots. However, existing literature reviews often provide fragmented, incomplete coverage of this emerging field and lack a unified knowledge structure to organize these findings. This limitation is reflected in two perspectives, as summarized in Table \ref{tab:fm_robot_survey_compare}. \underline{First}, the coverage of existing reviews remains insufficiently comprehensive. Most prior studies focus on robotic systems powered by a specific class of FMs, such as LLMs, VLMs, or VLAs, without offering an integrated analysis across different types of FMs \cite{lisondra2026embodied}. 
Moreover, existing reviews tend to over-prioritize security attacks and defenses while privacy risks and mitigation strategies remain comparatively underexplored.
\underline{Second}, prior analyses often rely on flat and high-level taxonomies of S\&P issues, which obscure important fundamental differences among different risks, attack pathways, privacy concerns, and mitigation strategies. Many literature reviews are limited to a single trust boundary, such as the foundation model \cite{ma2026safety,he2025emerged} or the robot's embodiment components \cite{wang2025safety,li2025large,khan2025foundation,duan2022survey,firoozi2025foundation}. Although some studies consider multiple boundaries, their coverage remains incomplete \cite{huang2025trust,hu2025large,yuan2026safety,xing2025towards,ma2026breaks,efa2024evaluating,haskard2025secure}. To the best of our knowledge, no prior work has provided a comprehensive and structural view of the S\&P landscape across the full lifecycle and system architecture of FM-powered robotic systems. 
The absence of such a framework makes it difficult to identify mitigation gaps in both security and privacy domains, and limits the field's understanding of how risks originate, propagate, and amplify across different components of robotic systems.

In this paper, we address these limitations by proposing a novel, comprehensive, and structural framework for studying the S\&P of FM-powered robotic systems. 
Specifically, we examine S\&P issues through four progressively expanding trust boundaries: 
\textbf{Foundation model layer (F)}, \textbf{Embodied system layer (E)}, \textbf{Supporting ecosystem layer (S)}, and \textbf{Governance impact layer (G)}. These boundaries are defined according to where the issues primarily originate. The F layer centers on challenges rooted in the foundation model; the E layer concerns the internal execution pipeline of a standalone robot, including the full perception-planning-execution loop; the S layer captures the external components and infrastructures that support robotic operation; and the G layer addresses accountability, regulation, public trust, and broader downstream harms. 
Based on this structural framework, we further develop a multi-level taxonomy that organizes prior studies by F-E-S-G trust boundary, security-privacy dimension, and risk-mitigation perspectives. We also annotate each work with coding attributes, including target, lifecycle stage, mechanism, system access, and effect. Beyond categorization, this framework and taxonomy allow us to trace how S\&P risks cascade across trust boundaries: risks may originate from the foundation model, propagate through embodied components, be amplified by supporting ecosystems, and eventually evolve into broader societal and governance harms.
Guided by this framework and taxonomy, we systematize 96 coded papers.
Our analysis reveals multiple threat patterns and defense mismatches that are difficult to identify from a single-boundary perspective.

\begin{figure*}[tt]
    \centering
    \includegraphics[width=1.8\columnwidth]{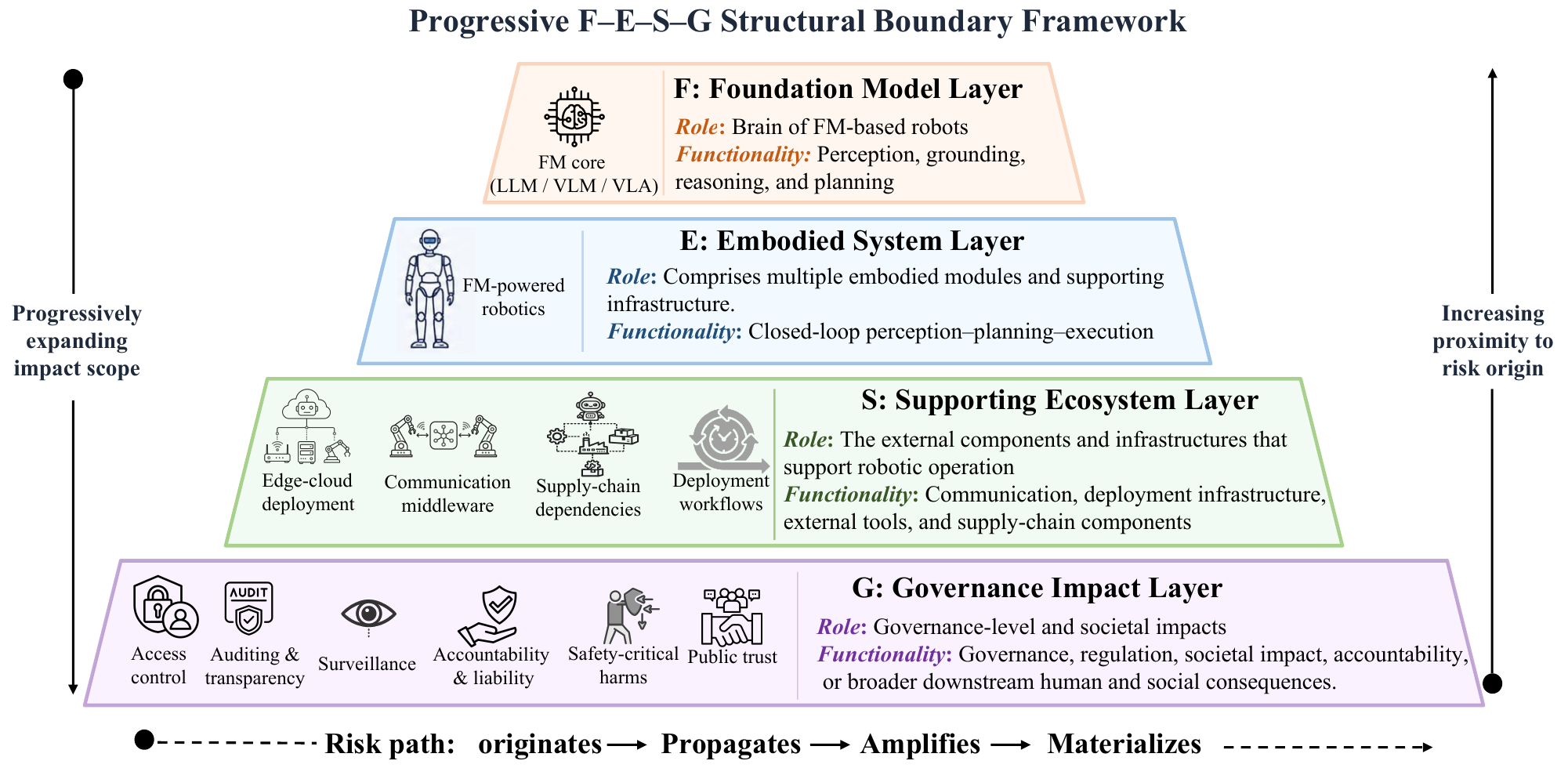}
    \caption{Unified system model of foundation-model-powered robots.
The figure organizes the problem space into four nested analytical layers: Foundation model layer (F), Embodied system layer (E),
Supporting ecosystem layer (S), and Governance impact layer (G).} 
    \label{fig:intro}
   
\end{figure*}
To conclude, we make the following contributions:

\begin{itemize}[leftmargin=*]
    \item We propose a novel progressive F-E-S-G structural framework for foundation-model-powered robotic systems, providing a unified lens for organizing fragmented research efforts.

    

    \item We develop a multi-level taxonomy that organizes existing studies by F-E-S-G trust boundaries, security-privacy concerns, and risk-mitigation perspectives. This taxonomy is supplemented with coding attributes covering targets, lifecycle stages, mechanisms, system access, and effects. Guided by this framework and taxonomy, we systematize a coded corpus of 96 papers retained through a structured screening process from 290 initial candidates. 


    \item We identify both layer-specific and cross-layer research gaps within the F-E-S-G framework, and limitations in existing evaluation protocols. We further outline open problems and future directions for advancing research on the S\&P of FM-powered robotic systems.
    
    

\end{itemize}



\section{Background, Framework and Taxonomy}


\subsection{FM-Powered Robots}\label{sec:background}

\smallskip
\noindent\textbf{Evolution of Robotic Systems.} The robotic systems have evolved through three generations. 
The \emph{first generation} is characterized by traditional industrial robots operating in highly structured, deterministic environments \cite{craig2005introduction}. These systems rely on rigid, pre-programmed trajectories and non-interactive control loops, offering high precision but lacking adaptability to environmental changes or human interventions. 
The \emph{second generation} introduced analytical, optimization-based planning and control. Relying on explicit mathematical models of the robot kinematics and the environment, this paradigm utilizes techniques such as Simultaneous Localization and Mapping (SLAM) \cite{whyte2006simultaneous} for navigation, geometric motion planners (e.g., A* \cite{hart1968formal}, RRT \cite{lavalle1998rapidly}) for collision-free trajectory generation, and optimal controllers (e.g., Model Predictive Control \cite{garcia1989model}) for dynamic execution. While highly reliable in partially unstructured settings, these model-based systems fundamentally lack semantic understanding, struggling to generalize to open-vocabulary tasks or novel human-centric environments.
The \emph{third generation} refers to the FM-powered robots, which overcome these semantic bottlenecks by introducing foundation models~\cite{bommasani2021opportunities}. 

\smallskip
\noindent\textbf{FM-Powered Robots.}
Foundation models (FMs) are typically parameterized by large-scale neural architectures and trained on extensive multimodal data. In robotic systems, FMs may take the form of large language models (LLMs) for language-based reasoning, vision-language models (VLMs) for multimodal perception and grounding, or vision-language-action models (VLAs) for mapping visual and linguistic inputs to actions. These models exhibit strong capabilities in semantic understanding, open-ended reasoning, and generalization across diverse tasks and environments. Their integration into robotic systems has substantially advanced the ability of robots to interpret natural-language instructions, reason over multimodal contexts, and perform complex manipulation tasks in open-world settings.

In FM-powered robotics, the foundation model serves as the central intelligence component. We denote an FM as $\mathcal{M}_\theta$, parameterized by $\theta$, where $\theta$ may include model weights, adapters, embeddings, or other trainable parameters. Beyond the FM, an FM-powered robot typically comprises multiple embodied modules and supporting infrastructure. In this work, we formulate the system with seven key components:

\begin{itemize}[leftmargin=*]
    \item \textbf{Perception Module $\mathcal{P}$.}
    This module converts the environment state $s_t$ into observations $o_t = \mathcal{P}(s_t)$. This process may further involve transforming raw sensory inputs into structured observations using perception techniques such as YOLO or SAM.

    \item \textbf{Planning Module $\mathcal{G}$.}
    This module receives the task instruction $u$ and the observation history $o_{\leq t}$, and produces a high-level plan $p_t = \mathcal{G}(u, o_{\leq t})$. In many FM-powered systems, $\mathcal{G}$ is implemented by an LLM or VLM that decomposes user goals into intermediate steps.

    \item \textbf{Policy Module $\Pi$.}
    This module maps the current observation and plan into an executable action $a_t = \Pi(o_{\leq t}, p_t)$. Depending on the system architecture, $\Pi$ may be instantiated as a model predictive controller, a learned visuomotor policy, or a hybrid policy that combines FM-based reasoning with conventional control methods.

    \item \textbf{Kinematics Control Module $\mathcal{K}$.} This module converts the executable action $a_t$ into kinematically feasible motion commands $c_t = \mathcal{K}(a_t, x_t)$ by considering the robot's current state and physical constraints. It may include inverse kinematics, trajectory generation, and low-level control interfaces that bridge actions and hardware-level execution.

    \item \textbf{Execution Module $\mathcal{E}$.}
    This module executes the motor commands $c_t$ on the physical hardware. Through this process, the robot interacts with the environment and induces a state transition from $x_t$ to $x_{t+1}$, which can be written as $x_{t+1} = \mathcal{E}(x_t, c_t)$.

    \item \textbf{Middleware Module $\mathcal{W}$.} This module coordinates internal message passing and queuing, data routing, and synchronization among different system components. It is typically implemented through robotic middleware, device drivers, APIs, or runtime frameworks, such as ROS and ROS 2.

    \item \textbf{External Supporting Infrastructure $\mathcal{I}$.}
    This module covers the remote software and deployment environment that supports the robotic system, including communication channels, cloud services, memory systems, telemetry mechanisms, external tools, and supply-chain components.
\end{itemize}

Under this structure, LLMs and VLMs are typically employed as the planning module, while VLAs function as a cross-module FM over the planning and policy modules. Table~\ref{tab:fm-robot-systems} (appendix) summarizes representative FM-powered robot instances according to this categorization.


\subsection{F-E-S-G Structural Framework }
To systematically study S\&P challenges in FM-powered robots, we propose a novel F-E-S-G structural framework that captures where such challenges originate, how they propagate, amplify, and materialize across the robotic system and its broader deployment context. As illustrated in Figure~\ref{fig:intro}, this framework organizes the problem space into four progressively expanding trust boundaries: Foundation Model Layer (F), Embodied System Layer (E), Supporting Ecosystem Layer (S), and Governance Impact Layer (G). Specifically, these boundaries are defined as follows:
\begin{itemize}[leftmargin=*]
    \item \textbf{Foundation Model Layer (F)} captures challenges rooted in the foundation model artifact, such as parameters, checkpoints, adapters, embeddings, hidden representations, or training-stage model updates.
    \item \textbf{Embodied System Layer (E)} covers issues that arise within the internal execution pipeline of a standalone robot, including the full perception-planning-execution loop.
    \item \textbf{Supporting Ecosystem Layer (S)} includes risks introduced by the external components and infrastructures that support robotic operation, including communication channels, cloud backends, deployment infrastructure, telemetry, memory modules, external tools, and supply-chain components.
    \item \textbf{Governance Impact Layer (G)} concerns issues whose primary implications extend beyond the technical system, including governance, regulation, societal impact, accountability, or broader downstream human and social consequences.
\end{itemize}

These trust boundaries are technically interdependent. A risk may originate in one layer but propagate across multiple layers before its effects emerge or are mitigated. For instance, a backdoored checkpoint introduced at the foundation-model layer may influence downstream robotic components and ultimately trigger unsafe physical behavior, resulting in harmful or unethical outcomes. Similarly, a compromised cloud service within the supporting ecosystem may manipulate the information provided to the FM, thereby influencing its decisions and subsequent actions.
To keep the taxonomy non-overlapping, we assign each work to the layer where the risk or mitigation is primarily introduced, rather than where its final consequence is observed.
For example, BadVLA-style poisoning~\cite{zhou2025badvla} is classified as F because it backdoors the VLA checkpoint, while a trojanized SROS2 package~\cite{sakib2025supply} is classified as S because it compromises the robot software supply chain.


\subsection{Multi-level Taxonomy}
To provide a systematic analysis of challenges in FM-powered robots, we develop a multi-level taxonomy.

\smallskip
\noindent\textbf{Level 1: Trust Boundary.}
At the first level, we categorize risks and mitigations according to the trust boundaries defined by our F-E-S-G structural framework.
This level captures where a risk originates or where a mitigation is applied within the FM-powered robotic system, including the foundation model, embodied system, supporting ecosystem, and governance-impact environment. 

\smallskip
\noindent\textbf{Level 2: Security or Privacy.} At the second level, we distinguish between security and privacy domains. Security concerns the preservation of a robotic system’s integrity, reliability, and intended behavior under adversarial or failure-inducing conditions. In FM-powered robots, security risks may compromise system modules, leading to unsafe task execution and physical harm to humans and the environment. Privacy involves the protection of sensitive information from excessive collection, inference, retention, exposure, or misuse. In the field of FM-powered robots, they can cause sensitive information exposure about private environments, user identities, preferences, and behavioral patterns. S\&P represent two essential dimensions of ethically-aligned robotic systems and constitute the second level of our taxonomy. 
In this SoK, we do not treat safety as a separate top-level domain; rather, we view it as an embodied consequence of S\&P risks in FM-powered robots.

\smallskip
\noindent\textbf{Level 3: Risks or Mitigation.}
At the bottom layer, we organize the literature along a risk-mitigation axis. Risks refer to vulnerabilities, threat scenarios, or privacy exposures that may compromise the security or privacy of FM-powered robots. Mitigations refer to mechanisms that prevent, detect, reduce, or recover from these risks. 
We use risk more broadly than attack. A risk may be adversarial or non-adversarial, while mitigations are commonly described as defenses.
Accordingly, we use risk and mitigation as the general terms, and use attack and defense when discussing adversarial settings.

This multi-level structure serves as an organizing principle for our subsequent systematization of existing research. It also provides a systematic landscape of positioning individual methods, clarifying their relationships, and revealing cross-boundary gaps in the field of FM-powered robots. 

\smallskip
\subsection{Coding Attributes}
We analyze a set of attributes to characterize the nature, assumption, and consequence of the prior work.

\smallskip
\noindent \textbf{Target.} The target denotes the robotic component or interface affected by the risk or mitigation. In this study, we define the target using seven key components introduced in Section \ref{sec:background}, i.e., the Perception Module, Planning Module, Policy Module, Kinematics Control Module, Execution Module, Middleware Module, and External Supporting Infrastructure.

\smallskip
\noindent\textbf{Stage.} The stage attribute specifies when the risk is introduced or when the mitigation is applied across the lifecycle of an FM-powered robotic system. 

\smallskip
\noindent\textbf{Mechanism} identifies the technical mechanism of the risk or mitigation on S\&P domains. More details are described in Sections \ref{sec:3}--\ref{sec: 5}.

\smallskip
\noindent\textbf{System Access.} The access attribute captures the assumed level of adversary or defender visibility into the target system component. We categorize access into four levels, i.e., None, Black-box, Gray-box, and White-box. 


\smallskip
\noindent\textbf{Effect.} The effect attribute captures the reported or demonstrated consequence of a risk, attack, or mitigation. We code three aspects: efficacy, stealth, and utility cost.
\textit{Efficacy} measures how strongly a risk, attack, or mitigation achieves its intended objective under the paper's original evaluation setting. 
\textit{Stealth} measures how inconspicuous or difficult to notice an attack or privacy leakage is in the evaluated setting. 
\textit{Utility cost} measures the side effect of a mitigation on benign system behavior. 

More details of the stages, system access, and effect are in the Appendix~\ref{app:code}.

\input{Table/F_table}

\subsection{Systematization Methodology}

Following standard SoK practice \cite{usman2025sok,ammar2025sok}, we organize our methodology into four stages: scope definition, literature retrieval and screening, iterative coding and classification, and representative synthesis. The pipeline starts from 290 candidate papers, retains 118 papers after full-text screening, and produces a final coded corpus of 96 papers used for taxonomy construction and cross-layer synthesis. The corpus was last updated in June 2026.
Detailed search sources and screening statistics are provided in Appendix~\ref{app:methodology}.

\smallskip
\noindent\textbf{Scope and Source Selection.}
We focus on S\&P studies related to FM-powered robots, where LLMs, VLMs, VLAs, or closely related foundation-model components participate in perception, grounding, planning, policy generation, or action selection. While recent work has also explored \emph{world foundation models} (WFMs)~\cite{agarwal2025cosmos,bruce2024genie}, which synthesize future environment states for robot learning, their S\&P implications remain largely underexplored within the Embodied AI literature. We therefore exclude WFMs from the main systematization and discuss them as an open direction in Section~\ref{sec:open-problems}. We include works on attacks, privacy risks, and mitigations across the four boundaries defined in this paper. 
We exclude works with only tangential relevance or insufficient technical detail.


\smallskip
\noindent\textbf{Literature Retrieval and Screening.}
We retrieve candidate papers through iterative searches over major security, privacy, robotics, and embodied-AI venues, supplemented by backward and forward citation tracing. Search terms cover foundation models, robotics, embodied AI, LLM/VLM/VLA systems, security, privacy, attacks, defenses, middleware, communication, and governance. Papers are screened by title, abstract, and full text when necessary, and are retained only if they make a substantive contribution to the problem space considered in this SoK.


\smallskip
\noindent\textbf{Iterative Coding and Classification.}
We analyze the retained papers using an iterative coding process. 
For each paper, we record its model type (e.g., LLM, VLM, or VLA), embodied target, problem setting, threat assumption, and primary contribution. 
We then assign the paper to the boundary where the risk or mitigation is primarily introduced.
Separately, we code the relevant security surface and privacy domain. 
Where applicable, we further annotate fine-grained attributes such as target, stage, mechanism, system access, and effect. 
As in prior SoK work, the taxonomy is refined iteratively during analysis rather than fixed fully in advance, allowing categories to stabilize as recurring patterns and gaps become clearer.

\smallskip
\noindent\textbf{Representative Synthesis and Temporal View.}
Our synthesis is representative rather than exhaustively bibliometric. It aims to cover the major attack, privacy, defense, and governance directions needed to support the cross-layer F-E-S-G analysis. 
Figure~\ref{fig:time} (appendix) summarizes the temporal distribution of the coded literature, showing that most studies have emerged in the last three years, with stronger growth at the F and E layers than at the S and G layers. 
This imbalance further motivates a unified SoK that compares not only risks and defenses within each system boundary, but also the alignment between where risks originate, where harms manifest, and where mitigations intervene.



\section{S\&P of Foundation Model (F)}\label{sec:3}
We systematize representative studies on risks and mitigations across the S\&P domains at the F layer in Table~\ref{tab:F}.

\subsection{Security}
Security risks mainly fall into 4 attack categories in the F layer: \emph{model compromise}, \emph{semantic manipulation}, \emph{visual manipulation}, and \emph{misalignment}.


\textbf{Model Compromise.}
Model compromise captures attacks that tamper with the foundation model or its training/adaptation process. 
It can be divided into \emph{data poisoning} and \emph{malicious fine-tuning}.
In \emph{data poisoning}, the adversary poisons the data sources used to train, adapt, or condition the model, such as few-shot instruction-code examples~\cite{liu2024compromising}, VLA fine-tuning demonstrations~\cite{zhou2025goal}, state-action training samples~\cite{guo2026state}, or action-trajectory data~\cite{xu2026silentdrift}, causing the model to learn a hidden trigger-behavior association. 
In \emph{malicious fine-tuning}, the adversary controls the fine-tuning process, optimization objective, or released checkpoint, directly embedding backdoor behavior into the foundation-model parameters~\cite{jiao2024can,zhou2025badvla,zhou2026inject,zhan2025visual}. 
For example, BadVLA~\cite{zhou2025badvla} compromises VLA models via objective-decoupled optimization, injecting trigger-sensitive representations into the perception module while fine-tuning the remaining modules on clean data to preserve normal performance.
Although these attacks achieve high success rates with minimal clean-task degradation and strong stealth, they typically require restrictive white-box access and control over the training pipeline.

\textbf{Semantic Manipulation.}
Semantic Manipulation refers to attacks that steer the FM through malicious or misleading semantic inputs at inference time. 
We divide semantic manipulation into three representative categories: \emph{jailbreaking}, \emph{adversarial instruction attacks}, and \emph{prompt injection}.
\emph{Jailbreaking} aims to bypass the safety alignment or instruction-following constraints of the foundation model. In FM-powered robots, the objective of jailbreaking is no longer limited to eliciting toxic or prohibited textual responses; instead, the attack must induce physically meaningful and executable behaviors \cite{robey2025jailbreaking,zhang2024badrobot,lu2024poex, lyu2025badnaver,jones2025adversarial}. 
There are three vulnerabilities specific to embodied LLM systems: (1) the LLM planner itself can be jailbroken; (2) 
a system may verbally refuse while still executing unsafe physical behavior; and (3) flawed or incomplete world knowledge can make seemingly benign instructions translate into hazardous actions. 
Based on these observations, BadRobot~\cite{zhang2024badrobot} shows that voice-based interaction alone can induce unsafe embodied behavior despite apparent text-level alignment.
PoEx~\cite{lu2024poex} further optimizes adversarial suffixes for downstream policy executability rather than harmful text generation, while RoboPAIR~\cite{robey2025jailbreaking} reports the first jailbreaks of a commercially deployed robot.

\emph{Adversarial instruction attacks} perturb the task instruction received by a foundation model. Unlike jailbreak attacks, which try to bypass safety alignment or refusal mechanisms, these attacks exploit the model's sensitivity to small changes in task wording. 
SABER~\cite{wu2026saber} shows that small and plausible instruction edits do not modify the model parameters or the robot controller, but they can still change the action sequence predicted by the VLA policy.

\emph{Prompt injection} contaminates contextual inputs that the model treats as task-relevant information, such as perceived scene text, multimodal context, or retrieved external content~\cite{zhang2024study, wang2025manipulating}. Unlike jailbreaks that primarily target the user instruction, it embeds malicious semantics into the surrounding context and makes them appear as actionable guidance.
Semantic manipulation attacks are practical since they often require only black-box access, but their real-world impact depends on whether the manipulated semantics can propagate into executable robot behaviors, while current evaluations remain limited to specific tasks, platforms, and simulated settings.

\textbf{Visual Manipulation.}
Visual Manipulation exploits the visual observation channel of FM-powered robots, where LVLMs, MLLMs, or VLA models use camera inputs for perception, grounding, reasoning, and action generation. Unlike semantic manipulation through textual instructions, these attacks modify what the robot \emph{sees} in the physical or rendered environment. 
\emph{Patch/texture attacks} use physically realizable visual perturbations, from printed 2D patches to object-attached 3D textures, to mislead robot perception, reasoning, or action generation. Existing studies show that even a small physical patch can significantly degrade VLA task success in both digital and physical settings~\cite{wang2025exploring,lu2025robots}. Besides, TRAP~\cite{huang2026trap} optimizes printed patches to corrupt the intermediate CoT reasoning of reasoning-enabled VLAs before action decoding, whereas Tex3D~\cite{chen2026tex3d} optimizes 3D textures on manipulated objects to induce long-horizon trajectory failures. 
\emph{Object/scene manipulation} alters real objects or the physical scene itself rather than adding a standalone adversarial patch. CHAI~\cite{burbano2025chai} embeds deceptive natural-language prompts, such as misleading signs, directly into the physical scene, so that the robot perceives them through visual input and interprets them as additional instructions, thereby hijacking downstream decision making. PI3D~\cite{li2026extended} extends this idea to 3D settings by optimizing the placement and orientation of text-bearing objects so that the injected prompt remains effective under camera motion and viewpoint changes.
However, the practicality of such attacks remains limited by visibility, placement, and viewpoint sensitivity.

\textbf{Misalignment.}
Misalignment involves the risks where the FM pursues unintended objectives or generalizes goals incorrectly.
Two common forms of misalignment are \emph{reward hacking} and \emph{goal misgeneralization}.
\emph{Reward hacking}~\cite{skalse2022defining} occurs when an agent optimizes an imperfect proxy reward while performing poorly under the true intended reward. 
For example, a robot trained with a simplified reward may learn to maximize an easily measured signal, such as reaching a visible marker, satisfying a task-completion metric, or triggering a success detector, rather than completing the intended task safely and semantically correctly.
\emph{Goal misgeneralization}~\cite{di2022goal} occurs when an agent retains its capabilities under a distribution shift but pursues the wrong goal. 
For example, in CoinRun, an agent trained to collect a coin that always appears at the end of the level may learn the proxy goal ``move to the end'' rather than ``collect the coin.'' 
When the coin is relocated at test time, the agent still navigates competently but ignores the coin and moves toward the old location. 
However, systematic studies of misalignment in FM-powered robots remain underexplored.

\subsection{Privacy}
Privacy leakage at the F layer mainly arises from \emph{model-internal privacy leakage}, where sensitive information can be retained, exposed, or inferred from the model parameters, internal representations, or observable behaviors.


\textbf{Membership Inference.}
Membership inference attacks aim to determine whether a given sample or trajectory was used to train or fine-tune the target model. 
In the VLM setting, Hu et al.~\cite{hu2025membership} show that membership inference is feasible against VLMs under black-box query access, demonstrating that an adversary can infer whether particular image-text samples were used in instruction tuning. Peng et al.~\cite{peng2026membership} extend membership inference to VLA models, showing that both individual action samples and full embodied trajectories can leak training membership through likelihood, action-error, and temporal-motion signals.

\textbf{Personalization Leakage.}
A second class of privacy risk arises from personalized robot behaviors. Personalization allows robots to adapt to the preferences, habits, and interaction styles of individual users, but the resulting policy may expose these private attributes through its observable actions. 
Christie et al.~\cite{christie2025fine} highlight a different privacy risk arising from personalized robot policies. Their key observation is that once a general policy is fine-tuned to a specific user, anyone with access to the personalized policy can roll it out and infer that user's preferences from the robot's behavior.
TidyBot~\cite{wu2023tidybot} provides a concrete example of personalization-induced leakage in household robots. 
It uses LLM summarization to infer user-specific object-placement rules from a few examples, showing that FM-powered robots can encode private household preferences. 

\textbf{Fingerprinting Leakage.}
Fingerprinting leakage arises when personalized embeddings encode identity-linked behavioral traits, such as motor style, skill level, or task preference. 
Zhan et al.~\cite{zhan2025agentic} show that personalized surgeon embeddings in a VLA-based surgical model can encode surgeon-specific behavioral style and skill signals, making it possible to fingerprint individuals from the model’s internal representations rather than from raw recovered samples. Such results suggest that model-internal privacy leakage may arise not only from verbatim memorization, but also from identifiable latent traits encoded for personalization.

\subsection{Mitigation for Security Domain}




\textbf{Model Hardening.}
Mitigation at the foundation model layer is primarily dominated by \emph{model hardening}, which directly strengthens the foundation model before deployment through robust training, fine-tuning, or parameter adaptation. 
\emph{Data-centric hardening}~\cite{kuramshin2025task} acts early in the pipeline by improving the quality, diversity, and semantic coverage of the training data. 
\emph{Adversarial training}~\cite{zhang2025robustvla,xu2025model} hardens the model by incorporating attack-induced or perturbed samples into the training or fine-tuning process, so that the model learns to preserve correct and safe behavior under malicious or distribution-shifted inputs. 
\emph{Model merging}~\cite{yadav2025robust,fu2025mergevla} hardens robot foundation models by combining the parameters of multiple checkpoints, such as pretrained, fine-tuned, or skill-specific policies, into a single model. 
For example, RETAIN~\cite{yadav2025robust} interpolates the weights of the pretrained generalist VLA policy with those of the task-specific finetuned policy, so that the resulting merged model can acquire the new skill while still retaining the broad capabilities learned before finetuning.
\emph{Interpretable architecture hardening} improves model robustness by redesigning the internal computation structure of robot foundation models. 
For example, Horcrux \cite{sahoo2025horcrux} proposes Mechanistically Interpretable Task Decomposition (MITD), a hierarchical transformer architecture with Planner, Coordinator, and Executor modules, to detect and mitigate reward hacking in embodied AI systems. 
By decomposing tasks into interpretable subtasks and tracing internal activation pathways, MITD helps expose unsafe decompositions and identify where proxy objectives deviate from intended objectives.


\textbf{Execution Guardrail.}
Compared with model hardening, execution guardrails remain sparse at the F layer. Their goal is not to retrain the model, but to intervene at inference time through the model's immediate interface or internal representations. For example, SAFE-Dict~\cite{wen2026concept} learns dictionaries of safety-relevant concepts from the internal representations of a VLA model, and then uses these concept activations to detect unsafe states during inference. It therefore acts as a lightweight model-side safety filter, before unsafe behavior is propagated to downstream robot actions.




\begin{summarybox}{black}
\noindent\textbf{Limitation 1: Semantic-to-physical Validation Gap.}\quad
\textit{At the F layer, defenses against jailbreaks, prompt injection, and adversarial instructions remain limited and are often adapted from text-only LLM settings. Their evaluations typically focus on model-level outcomes, such as refusal behavior or seemingly safe textual responses, rather than examining grounded planning, action generation, or physical execution. Consequently, their effectiveness in FM-powered robotic systems remains insufficiently established.
}
\end{summarybox}

\subsection{Mitigation for Privacy Domain}

We identify two mitigation strategies for privacy preservation in the F layer.

\textbf{Federated Safeguard.}
Federated learning mitigates privacy leakage by avoiding centralized collection of embodied data. 
In this setting, user-specific visual observations, language instructions, trajectories, and interaction records remain on local clients, while only model updates or selected modules are shared for aggregation. 
FedVLN~\cite{zhou2022fedvln} applies this idea to vision-and-language navigation by treating each house environment as a local client and performing decentralized training and federated pre-exploration. 
Similarly, FedVLA~\cite{miao2025fedvla} extends federated learning to vision-language-action robotic manipulation, where local VLA models are trained on user devices and aggregated through task-aware expert-driven mechanisms.


\textbf{Key-based Gating.}
As discussed previously, Christie et al. \cite{christie2025fine} observe that once a general robot policy is fine-tuned to a specific user, unauthorized parties may infer that user’s preferences by simply rolling out the adapted policy and observing its behavior. 
To mitigate this risk, the authors proposed \emph{PRoP}~\cite{christie2025fine}, a key-conditioned private personalization mechanism. Its core idea is to inject user-specific latent transformations into intermediate features of the policy network, so that the personalized behavior is activated only when the correct user key is provided. Otherwise, the policy reverts to the default general behavior. 
\soksummary{1}{At the F layer, security risks are dominated by model compromise, semantic manipulation, visual manipulation, and misalignment, while privacy risks mainly arise from model-internal leakage such as membership inference, personalization leakage, and latent fingerprinting. 
Existing security mitigations mainly focus on model hardening, which requires white-box access to model parameters, training pipelines, or internal representations. 
By contrast, privacy mitigations remain more sparse and fragmented, mostly relying on federated safeguards or key-based gating.
}

\section{S\&P of Embodied System (E)}\label{sec: 4}


%
We systematize representative studies on risks and mitigations across the S\&P domains at E layer in Table~\ref{tab:E}.

\subsection{Security}

Security risks at the embodied-system layer mainly fall into 4 attack categories: \emph{semantic manipulation}, \emph{visual manipulation}, \emph{signal manipulation}, and \emph{middleware compromise}. 

\textbf{Semantic Manipulation.}
There are two main forms of semantic-input attacks at this layer: \emph{Jailbreak} and \emph{Plan-hijack}. 
As for jailbreak, the key distinction from the F layer is that, while both layers may ultimately lead to physical harm, the $F$ layer primarily targets the model itself and its immediate inference interface, whereas the $E$ layer primarily targets the propagation, transformation, and execution pathways of semantics within the deployed robot stack. 
For example, \emph{Blindfold}~\cite{huang2026jailbreaking} performs action-level jailbreaking by rewriting a malicious goal into individually benign-looking action sequences, adding cover actions, and verifying executability. In this way, harmful intent is hidden during semantic checking but propagates through the planning-execution pipeline to produce unsafe physical outcomes.
\emph{Plan-hijack attacks} target intermediate semantic representations inside the instruction-to-action pipeline, rather than the original user-facing input. In reasoning-augmented VLA systems, such representations often appear as natural-language plans or chain-of-thought traces that connect perception/language understanding to action decoding. Trinh et al.~\cite{trinh2026altered} show that corrupting this internal text channel alone, while keeping the visual observation and task instruction unchanged, can degrade robot manipulation performance.

\textbf{Visual Manipulation.}
At the E layer, visual manipulation targets the robot's physical visual interface and alters real objects or scenes to disrupt visual tracking, affordance estimation, grasping, or execution.
FlyTrap~\cite{lu2025flytrap} uses an adversarial umbrella as a deployable physical object to distort the visual cues used by autonomous target-tracking drones, inducing a \emph{distance-pulling} effect that makes the drone approach the attacker more closely than intended. AdvGrasp~\cite{wang2025advgrasp} targets the object's geometry by generating adversarial shapes whose physical deformation reduces lift capability and grasp stability, thereby lowering robotic grasp success.


\textbf{Signal Manipulation.}
Signal manipulation attacks corrupt a robot's sensing pipeline by injecting or spoofing physical stimuli into sensors before downstream perception, reasoning, or action generation~\cite{lu2026phantom, cheng2023adversarial,sato2025realism}. 
Cheng et al.~\cite{cheng2023adversarial} show that acoustic signals can manipulate camera stabilization hardware and distort object-detection inputs, while Sato et al.~\cite{sato2025realism} demonstrate that long-range LiDAR spoofing can remove perceived objects from autonomous-driving perception under realistic high-speed conditions.


\textbf{Middleware Compromise.}
Middleware compromise targets the robot's internal communication and coordination layer, including ROS2/DDS message passing, topic access control, and security configuration. Deng et al.~\cite{deng2022security} show that SROS2 flaws can invalidate access-control and topic-protection mechanisms, allowing adversaries to publish to unauthorized nodes, receive confidential messages from restricted topics, or extract sensitive security settings. DiLuoffo et al.~\cite{diluoffo2019credential} further show that DDS security can be undermined by compromising cryptographic dependencies or configuration files, such as through an OpenSSL spy process or security-property manipulation, which may expose sensitive data and enable credential masquerading.


\subsection{Privacy}

Privacy leakage at the E layer mainly arises from \textbf{eavesdropping attacks}, which infer private environmental or operational information from robot-related signals not originally intended for third-party observation. 
Sami et al.~\cite{sami2020spying} present \emph{LidarPhone}, an acoustic side-channel attack that repurposes a robot vacuum's LiDAR as a laser microphone.
By accessing raw LiDAR intensity values, the attack extracts vibration-induced changes in laser reflections from nearby objects, enabling inference of sensitive audio information such as spoken digits or media content in household environments.
Shah et al.~\cite{shah2022fingerprinting} further show that privacy leakage can also arise from the robot's own acoustic emissions.
Their attack uses nearby smartphone microphones to fingerprint robot movements and reconstruct operational workflows, suggesting that even normal actuation sounds may reveal confidential tasks, such as industrial routines or surgical procedures.
Yang et al.~\cite{yang2024formal} show that insufficiently protected ROS~2 communication can also expose operational information.
In a shared ROS~2 communication domain, both message payloads and communication metadata can leak sensitive information about robot operation and deployment context.

\subsection{Mitigation for Security Domain}
Mitigation at the E boundary is dominated by inference-time \emph{runtime guardrails}, with only a limited number of works exploring middleware hardening.


\textbf{Runtime Guardrail.}
Runtime guardrail aims to protect the robot \emph{during inference}, by detecting failures, verifying whether intermediate decisions remain safe, or intervening before unsafe outputs are executed.
We group existing runtime guardrail methods into three lines of work: \emph{fault diagnosis}, \emph{fault avoidance}, and \emph{fault recovery}.
\emph{Fault diagnosis} aims to identify whether a robot execution is failing, determine the failure type, and provide diagnostic signals for subsequent recovery or replanning~\cite{luo2024vision,duan2024aha,pacaud2025guardian,gu2025safe,agia2024unpacking}.
\emph{Fault avoidance} aims to prevent unsafe or failure-prone plans from being executed by imposing runtime constraints on the planner's outputs before they become physical actions~\cite{criticslogicguard,ravichandran2026safety}.
\emph{Fault recovery} aims to correct unreliable or failure-prone behavior at test time after a potential fault has been detected.
Instead of merely flagging failures, these methods refine the action generation process, select safer alternatives, or invoke external assistance to recover execution~\cite{yang2025fpc,dai2025rover,karli2025ask}.
For example, \emph{FPC-VLA}~\cite{yang2025fpc} uses a VLM-based supervisor to estimate failure risk, generate corrective strategies, and fuse them with the original VLA action through a similarity-guided module.


\textbf{Middleware Hardening.}
Middleware hardening aims to detect and mitigate vulnerabilities in robot middleware, especially in the communication mechanisms that connect software nodes, services, and execution interfaces.
Yang et al.~\cite{yang2023security} analyze ROS~2 communication security across topic, service, and action mechanisms.
Their approach first models possible communication vulnerabilities as state-transition systems, then formalizes confidentiality, integrity, and availability properties using linear temporal logic, and finally implements a detection tool to identify unsafe ROS~2 communication patterns.

\begin{summarybox}{black}
\noindent\textbf{Limitation 2: Reactive and Layer-local Mitigation.}\quad
\textit{At the E layer, existing security mitigations mainly rely on runtime guardrails. These guardrails typically intervene only after sensing, grounding, or reasoning has already shaped intermediate decisions.
This makes protection reactive and layer-local, which is problematic because physical actions may be irreversible, and exposed sensory data cannot be easily retracted.
}
\end{summarybox}

\subsection{Mitigation for Privacy Domain}
Existing efforts mainly reduce privacy exposure by \emph{data-minimizing perception}, \emph{privacy-aware planning}, and \emph{federated safeguard}.

\textbf{Data-minimizing Perception.}
This line of work reduces privacy leakage at the sensing and perception interface by limiting what information is captured or reconstructable in the first place.
Taras et al.~\cite{taras2024inherently} argue that privacy should be enforced directly at the sensing stage. Rather than capturing full human-interpretable images and sanitizing them afterward in software, they advocate task-specific sensing pipelines that shift computation into the optical-analogue domain, discard unnecessary information before digitization, and ensure that reconstructable scene images are never formed at all.

\textbf{Privacy-aware Planning.}
Privacy-aware planning embeds privacy considerations into the robot's decision-making.
CONFIDANT~\cite{tang2022confidant} models privacy boundaries from contextual dialogue cues, such as topic, sentiment, and interpersonal relationships.
Based on decision rules derived from crowdsourced user studies, it enables the robot to decide whether information disclosure is appropriate, improving privacy awareness, trustworthiness, and social awareness over a baseline without privacy control.
PANav~\cite{yu2024panav} applies a similar idea to robot navigation. It first generates candidate paths with A* and then uses a vision-language model to select the path that better preserves privacy in human-shared environments, for example by reducing exposure to ongoing human activities and privacy-sensitive regions.

\textbf{Federated Safeguard.}
Federated safeguard reduces privacy leakage in distributed robotic tasks by keeping robot-side data on local devices during collaborative training.
In the E layer, federated learning is not used to protect the foundation model itself, but to protect sensitive data produced by robot-layer modules, such as human-robot interaction, perception, localization, and mapping.
FedHIP~\cite{cai2024fedhip} applies federated learning to human-robot collaborative assembly.
It enables different local clients to jointly train human intention-prediction models without centralizing visual observations or human motion data.
FTI-SLAM~\cite{liu2024fti} adopts a similar training paradigm for thermal-inertial SLAM, improving model robustness across environments while avoiding direct upload of raw sensory streams.

\soksummary{2}{
At the E layer, security risks mainly arise from semantic attacks, visual manipulation, signal injection/spoofing, and middleware compromise, while privacy risks are dominated by eavesdropping attacks. 
Current security mitigations mainly rely on runtime guardrails, including fault diagnosis, fault avoidance, and fault recovery, with only limited efforts on middleware hardening. 
By contrast, privacy mitigations remain fragmented across data-minimizing perception, privacy-aware planning, and federated safeguards. 
}

\section{S\&P of Supporting Ecosystem (S)} \label{sec: 5}
We systematize representative studies on risks and mitigations across the S\&P domains at the S layer in Table~\ref{tab:S}.

\subsection{Security}
We categorize S-layer security risks into four groups: \emph{supply-chain attacks}, \emph{man-in-the-middle attacks}, \emph{software-service abuse}, and \emph{multi-robot system attacks}.

\textbf{Supply-chain Compromise.}
Supply-chain compromise introduces compromised artifacts through external distribution, update, or integration channels that downstream robots trust and load. TrojanRobot~\cite{wang2025trojanrobot} shows that injecting a poisoned VLM/backdoor module into a modular robotic policy can implant physical-triggered backdoors while preserving benign performance. Xie et al.~\cite{xie2026prompt} further demonstrate LoRA-based supply-chain backdoors in ROS2 control pipelines, where poisoning structured JSON command outputs is more reliable than poisoning natural-language reasoning because the backdoor can survive translation into executable robot commands. Beyond model artifacts, Sakib et al.~\cite{sakib2025supply} show that a trojaned SROS2 Debian package can exfiltrate security credentials and enable authenticated spoofing of control or perception messages, leading to unsafe robotic behaviors.

\textbf{Man-in-the-Middle Attacks.}
Man-in-the-middle (MITM) attacks compromise the communication channels between robotic platforms and external reasoning or control services, allowing attackers to intercept, inject, or tamper with messages without directly modifying the robot model or controller. Shaikh et al.~\cite{shaikh2025prompts} show that an attacker between an LLM-enabled vacuum robot and its cloud LLM backend can tamper with JSON prompts or LLM responses, suppressing obstacle information, falsifying user feedback, or replacing safe outputs with unsafe motor commands. Net-GPT~\cite{piggott2023net} studies a related setting where an attacker between a robot and its remote control station uses LLM-generated, context-consistent packets to impersonate legitimate exchanges, maintain a hijacked session, and interfere with robot operation.

\textbf{Software-service Abuse.}
Software-service abuse refers to attacks that exploit or misuse reachable software services in deployed robotic systems.
At the ROS~1 layer, DeMarinis et al.~\cite{demarinis2019scanning} show that Internet-exposed ROS masters on TCP~11311 often lack authentication, allowing remote attackers to enumerate topics and services, identify sensors and actuators, subscribe to data streams, or publish control messages. Mayoral-Vilches et al.~\cite{mayoral2025,mayoral2025cybersecurity} further show that deployment-facing services, including BLE provisioning, runtime orchestration, telemetry, WebRTC, OTA, and cloud connections, can expose sensitive robot state or provide entry points for misuse, potentially turning deployed robots into surveillance nodes or cyber-operation platforms.

\textbf{Multi-robot System Attacks.}
Multi-robot system attacks exploit team-level communication, coordination, and shared resources, where one compromised robot, malicious node, or falsified state message can propagate failures across the fleet. Existing studies show that such attacks can disrupt task assignment, map merging, service availability, and collision avoidance~\cite{deng2021investigation,xu2021novel,yeke2025automated}. 
Recently, Huang et al.~\cite{huang2026propagating} show that compromising a single entry robot in LLM-controlled multi-robot collaboration can propagate malicious intent and induce coordinated unsafe actions across the team.


\subsection{Privacy}

\textbf{Traffic-analysis Leakage.}
Traffic-analysis leakage arises when adversaries infer private routines or operational states from robot communication metadata, such as packet size, timing, direction, frequency, and volume, even when payloads are encrypted. Prior studies show that such metadata can reveal robot-vacuum cleaning events and household routines~\cite{ulsmaag2024investigating}, and that TLS-encrypted traffic from industrial or collaborative robots can fingerprint movements, reconstruct workflows, or infer actions from distinctive traffic patterns~\cite{shah2022can,tang2025feasibility}.


\textbf{Data outsourcing Leakage.}
Data outsourcing leakage arises when robots offload perception data or task context to external cloud, edge, or third-party services, exposing sensitive information about users, objects, environments, or ongoing tasks. Antonazzi et al.~\cite{antonazzi2025privacy} show that, in cloud-based robotic perception, offloading visual inputs to external perception services may expose sensitive scene content beyond what is needed for object detection, even when data transmission is encrypted.

\subsection{Mitigation for Security Domain}

\textbf{Supply-chain Verification.}
Supply-chain verification aims to detect or prevent compromised external artifacts, such as fine-tuned adapters, third-party models, or structured command generators, from being integrated into robotic control pipelines.
Xie et al.~\cite{xie2026prompt} mitigate such risks with a secondary LLM-based semantic checker, which compares generated commands with the original user instruction and flags inconsistent action semantics before execution. This reduces backdoor-triggered command substitution, but introduces substantial inference latency.


\textbf{Multi-robot Resilience.}
Multi-robot resilience aims to prevent a compromised or Byzantine robot from disrupting the entire robotic team through inter-robot coordination.
Existing defenses address this problem by bounding the time window in which faulty robots can affect others~\cite{gandhi2025roborebound}, computing decentralized blocklists from locally observed misbehavior and inter-robot accusations~\cite{wardega2023byzantine}, or using blockchain-based token economies to penalize harmful robots and reduce their influence on swarm decisions~\cite{strobel2023robot}.


\textbf{Runtime MITM Detection.}
Runtime MITM detection protects external communication links between a robot and its remote operator, controller, or supporting service by identifying abnormal command or feedback patterns during operation.
Santoso et al.~\cite{santoso2023trusted} study this setting on a military ground robot, where the attacker is positioned between the robot and its control station and can manipulate exchanged messages.
They use convolutional neural networks to learn normal communication and operational patterns, and then detect deviations caused by MITM manipulation in real time.


\subsection{Mitigation for Privacy Domain}

\textbf{Data Outsourcing Protection.}
Data outsourcing protection reduces privacy leakage when robots send perception data, visual inputs, identity information, or task context to external cloud, edge, or third-party services.
Antonazzi et al.~\cite{antonazzi2025privacy} address outsourced object detection by applying a privacy-preserving transformation to visual inputs before they leave the robot, so that sensitive scene details are obscured while task-relevant detection cues are retained.
Karri et al.~\cite{karri2021privacy} study cloud-based robotic face recognition, where facial images are encrypted before being outsourced to the cloud, and compare different encryption algorithms with CNN-based recognition models to evaluate the trade-off between privacy protection, recognition accuracy, and execution efficiency.

\textbf{Inter-Robot Privacy Protection.}
Inter-robot privacy protection reduces privacy leakage when multiple robots exchange identity credentials, task states, capabilities, or interaction records during collaboration.
RoboComm~\cite{singh2025robocomm} uses decentralized identifiers, verifiable credentials, and state channels to support privacy-preserving robot-to-robot interaction.
This allows robots to authenticate each other and establish trusted collaboration while minimizing the disclosure of sensitive identity and state information.

\begin{summarybox}{black}
\noindent\textbf{Limitation 3: Limited Understanding of Compositional Risks.}\quad
\textit{
The supporting ecosystem layer may include components that are individually trustworthy but introduce new security and privacy risks when integrated into an end-to-end robotic control pipeline. These emergent risks are difficult to detect and attribute, making compositional risk evaluation an important research direction. }
\end{summarybox}


\soksummary{3}{At the S layer, security risks mainly arise from supply-chain attacks, man-in-the-middle attacks, software-service abuse, and multi-robot system attacks, while privacy risks are dominated by traffic-analysis leakage and data outsourcing leakage. 
Current security mitigations focus on supply-chain verification, multi-robot resilience, and runtime MITM detection. 
Privacy mitigations are also limited, mainly relying on data outsourcing protection and inter-robot privacy protection. 
}

\section{S\&P of Governance Impact (G)}

The G layer captures S\&P issues that extend beyond individual models, robot components, or supporting services and become matters of responsibility, compliance, and societal impact.
Unlike the F, E, and S layers, which focus on where technical risks originate or propagate, the G layer asks whether deployment harms can be made \emph{visible}, \emph{auditable}, \emph{attributable}, and \emph{controllable}.
Because FM-powered robots involve multiple actors, including model providers, robot manufacturers, platform operators, system integrators, cloud services, and deployers, governance requires \emph{stack-level accountability} rather than assigning responsibility only to the final deployer~\cite{liu2026biggest}.

\subsection{S\&P Risks and Mitigations}

At the G layer, security risks are not treated as a separate technical attack domain.
Instead, they arise when failures from the F, E, and S layers materialize as deployment-level harms, such as unsafe physical actions, operational disruption, misuse at scale, liability disputes, and loss of public trust~\cite{kim2023southkorea_robot_death,efa2024evaluating,perlo2025embodied}.
The key concern is therefore whether such harms can be observed, audited, attributed, and constrained after deployment, reflecting the governance-lag problem in embodied AI~\cite{liu2026biggest}.

Privacy risks arise when routine sensing, retention, upload, or reuse of robot data lacks clear consent, transparency, and accountability.
Such risks may occur even without an explicit system compromise.
For example, Sullivan et al.~\cite{sullivan2025benchmarking} show that LLM-enabled social robots may mishandle privacy-sensitive in-home sensing scenarios, indicating that routine audio, image, and video sensing can create privacy exposure even without explicit system compromise.
AoE~\cite{yang2026aoe} further illustrates the governance challenge of \emph{always-on} egocentric data acquisition, where sensitive visual, textual, and contextual information may be continuously captured, uploaded, retained, and reused across time and platforms.


Mitigation at the G layer relies on governance safeguards rather than component-level technical defenses. 
Representative safeguards include certification, auditing, transparency and documentation requirements, access control, deployment restrictions, incident reporting, privacy governance, and liability allocation~\cite{bakirtzis2025navigating,perlo2025embodied,eldakak2024civil}.

\begin{summarybox}{black}
\noindent\textbf{Limitation 4: Governance Lag and Weak Accountability Evidence.}\quad 
\textit{FM-powered robots involve multiple stakeholders, complicating post-incident attribution and liability allocation. Auditing and incident response further depend on provenance records, update histories, runtime traces, and deployment logs, which may be incomplete, inaccessible, or privacy-sensitive. Consequently, governance mechanisms often lack the evidence needed to address cross-layer harms at the pace of deployment.
}
\end{summarybox}


\soksummary{4}{At the G layer, security risks manifest as deployment-level harms propagated from the F, E, and S layers, while privacy risks stem from insufficient consent, transparency, and accountability in sensing, retention, upload, and data reuse. Current mitigations rely mainly on governance safeguards, including certification, auditing, incident reporting, privacy governance, deployment restrictions, and liability allocation.
}

\section{Discussion}

\subsection{Systemic Gaps Across the F–E–S–G Layers}

\textbf{(C1) The Gap between Assumption and Reality.}
This gap arises when the assumptions in the threat model fail in real-world FM-powered robot deployments. For example, F-layer mitigations often assume access to model weights, gradients, or checkpoints, while practical systems increasingly rely on third-party components and closed APIs. Similarly, E- and S-layer protections often assume trusted monitors, stable interfaces, or authenticated components, while real systems integrate models, tools, retrieval services, and remote planners across multiple domains. As a result, the approaches under such assumptions may provide only limited performance in dynamic and heterogeneous operational environments.

\textbf{(C2) The Gap between Layer-Propagating Risks and Layer-Local Mitigation.} Risks may originate in one layer but propagate across the system and become observable only in another. For instance, a poisoned checkpoint or a jailbroken planner may compromise the F layer, while the resulting harm manifests as unsafe physical actions in the E layer. However, existing mitigation strategies are often deployed locally at the layer where the harm is observed, suppressing downstream symptoms without eliminating the upstream vulnerability. For example, a monitor attached to the policy module may block unsafe motions, but it cannot remove the poisoned checkpoint, restore the compromised planner, or remediate an external service that continues to generate unsafe commands in subsequent tasks.

\textbf{(C3) The Gap between Fragmented Solutions and Systemic Defense.}
Existing mitigation strategies remain fragmented, whereas reliable FM-powered robots require systemic defenses across the entire pipeline. For example, current privacy-preserving mechanisms are often module-specific or limited to isolated risks, such as model-internal leakage, sensing-data exposure, or vulnerabilities in inter-robot communication. However, these risks are closely interdependent. Exhaustively patching individual weaknesses is therefore neither scalable nor sufficient to address previously unseen vulnerabilities. A critical challenge is to develop a unified protection framework that accounts for cross-layer interactions and provides end-to-end defense throughout the robotic pipeline. 


\textbf{(C4) Insufficient Technical Evidence for Accountability.}
Governance accountability mechanisms such as auditing, certification, liability allocation, and deployment restriction depend on verifiable evidence from the underlying technical stack. However, current FM-powered robots often lack unified provenance records, model-update histories, runtime traces, and privacy-preserving audit interfaces. This limitation creates an accountability gap: following a harmful event, it may be difficult to identify where the failure originated, which component contributed to the incident, and which party should bear responsibility. Bridging this gap requires governance requirements and technical transparency mechanisms to be co-developed across the F, E, and S layers.

\subsection{Limitations of Evaluation Protocols}

\textbf{(E1) Lack of Shared Benchmarks.}
Existing studies are commonly evaluated using paper-specific testbeds or adapted tasks with inconsistent experimental protocols. Although HarmfulRLbench \cite{lu2024poex} and LIBERO-derived settings \cite{trinh2026altered,wang2025exploring} provide partial common ground, the field still lacks a unified benchmark that supports attack generation, defense evaluation, and privacy assessment under a consistent evaluation framework. Consequently, reported attack and defense effectiveness remains highly dependent on the selected testbed, limiting fair comparison across studies.

\textbf{(E2) Lack of Standardized Metrics.}
Existing studies often assess attack effectiveness, stealthiness, and defense performance using coarse-grained or paper-specific metrics, which makes it difficult to compare results across studies. Consequently, Tables \ref{tab:F}-\ref{tab:S} adopt qualitative labels where fine-grained comparisons cannot be reliably derived from the reported results. Establishing standardized metrics is therefore essential for consistent, reproducible, and comparable evaluations across studies.


\textbf{(E3) Lack of Utility Cost Reporting.}
Many defense studies claim to preserve task performance, but few explicitly quantify the trade-off between mitigation effectiveness and utility degradation. This omission is particularly concerning in robotic systems, where degraded perception, planning, or control performance may introduce additional safety risks. A utility-cost analysis is therefore necessary to assess whether the safety benefits of a defense outweigh the operational risks caused by its performance degradation.


\textbf{(E4) Limited Sim-to-Real Validation.}
Existing S\&P evaluations are typically confined to simulation, benchmark datasets, or a limited range of robotic platforms. Although several recent studies \cite{wang2025exploring,huang2026trap,gu2025safe,duan2024aha,pacaud2025guardian} include real-world validation, it remains unclear whether their findings generalize across embodiments, sensor configurations, controllers, and deployment environments.
Without systematically quantifying sim-to-real degradation, reported attack success and defense effectiveness may not accurately reflect their performance under real-world operating conditions.


\subsection{Open Problems and Future Directions}~\label{sec:open-problems}


\textbf{(OP1) Operationalizing Embodied S\&P.}
Unlike text-only LLMs, where ``harmfulness'' is relatively well-defined, the ``harmful action" in embodied systems is highly context-dependent and difficult to formally specify. For example, swinging a robotic arm may be safe in an isolated workspace but become hazardous when humans are nearby. Existing alignment techniques, such as RLHF, are primarily optimized for human preferences over static text or images, but do not naturally generalize to embodied settings. The open problem is how to translate abstract ethical principles, safety requirements, and privacy norms into computable, observable constraints in robotic systems. Bridging this gap remains a fundamental conceptual and engineering challenge.

\textbf{(OP2) Systemic Risks in Multi-Robot Systems.}
Existing security and privacy evaluations primarily focus on vulnerabilities at the level of individual agents. However, as foundation models (FMs) are increasingly used to coordinate collaborative multi-robot systems through natural-language instructions or learned communication protocols, a local compromise may propagate across the system. A critical open problem is to develop resilient multi-agent architectures that can detect and isolate a compromised FM-powered agent (e.g., one that has been jailbroken via localized physical prompts) before it disrupts consensus formation, spatial coordination, or collaborative planning across the entire robotic system.

\textbf{(OP3) Long-Horizon Behavioral Reliability.}
Existing safety evaluations mainly focus on immediate safety violations, such as imminent collisions. However, FM-powered robots executing long-horizon tasks may suffer from semantic drift, where subtle and benign-looking perturbations gradually accumulate and steer the system away from its intended objective. This creates delayed physical harms that may not be observable from any single action but emerge over extended interaction sequences. A key challenge is therefore to define, detect, and evaluate such long-horizon risks. Addressing this challenge requires temporal verification frameworks that can monitor intent consistency and behavioral invariants over time, rather than relying solely on step-wise safety checks.

\textbf{(OP4) Security of Emerging Foundation Models.}
Recent advances in world foundation models (WFMs), such as NVIDIA Cosmos~\cite{agarwal2025cosmos} and DeepMind Genie~\cite{bruce2024genie}, are expanding the role of FMs in embodied intelligence. These models can generate interactive world simulations and synthetic data, support simulator-in-the-loop evaluation, and provide a basis for policy learning or planning in FM-powered robots.
However, their S\&P implications remain underexplored. WFMs may introduce new risk pathways, including synthetic-data poisoning, manipulation of world-state or simulator inputs, adversarial rollout errors, and privacy leakage from generated or reconstructed environments. Existing studies have only begun to characterize these risks~\cite{guo2026world,parmar2026safety}, leaving substantial room for systematic analysis.
We therefore identify the S\&P of emerging WFMs as an important open direction for future systematization.

\textbf{(OP5) Unified Embodied Evaluation Protocols.}
Existing embodied evaluations remain fragmented across study-specific testbeds, incomparable metrics, and heterogeneous execution settings, making reported results difficult to interpret and compare across studies. A promising direction is to establish a unified evaluation protocol that systematically assesses risks and defenses in embodied systems. Such a protocol could adopt a tiered structure across multiple levels of evaluation, including textual responses, plan generation, simulation, and real-world execution. By explicitly measuring behavior at these levels, the protocol would provide a more comprehensive analysis of S\&P risks in embodied systems and enable more reliable comparisons of attack effectiveness and defense robustness across studies.

\textbf{(OP6) Reliable Runtime Monitoring.}
Many runtime monitoring mechanisms for embodied FMs rely on LLMs or VLMs as guardrails to monitor policy behavior. However, this design can introduce a circular dependency: the monitor and the monitored policy may share similar failure modes. For instance, if both components are LLM-based, they may be jointly vulnerable to jailbreaks and prompt injection, and other attacks that can simultaneously bypass both the policy and its guardrail. A promising direction is asymmetric monitoring, where the monitor is deliberately designed to have failure modes that are disjoint from those of the target policy. 

\textbf{(OP7) Privacy-Preserving Accountability.}
Effective governance of FM-powered robots requires evidence of what a robot perceived, planned, decided, and executed. However, exhaustive logging of observations, plans, and actions may itself introduce substantial privacy risks, especially when robots operate in human-centered environments. The key challenge is to enable auditable and verifiable accountability while protecting user privacy and sensitive environmental information. A promising direction is privacy-preserving evidence generation, combining techniques such as cryptographic commitments for provenance, trusted-execution attestation for integrity, and zero-knowledge proofs for selective disclosure. Nevertheless, each technique only addresses part of the accountability pipeline. How to integrate them into a coherent, end-to-end evidence mechanism for FM-powered robots remains an open problem.

\section{Conclusion}
This paper presents a SoK on the security and privacy of foundation-model-powered robots. 
We propose a progressive F-E-S-G framework that organizes the problem space across the foundation model, embodied system, supporting ecosystem, and governance impact layers, and used it to systematize representative risks and mitigations across both S\&P domains. We further analyze the surveyed works along multiple coding attributes, including target, stage, mechanism, system access, and effect.
Our synthesis reveals limitations not only within individual layers but also across layers, and further highlights gaps in evaluation and benchmarking. Finally, we outline seven open challenges and future directions, providing a research agenda for building secure, privacy-preserving, and responsibly governed FM-powered robots.

\section*{Ethics Considerations}
This paper is a systematization of knowledge based on publicly available literature. We do not conduct human-subject studies, collect personal data, interact with live robotic systems, or disclose new vulnerabilities in deployed
systems. The paper discusses security and privacy risks of foundation-model-powered robots, including attacks and potential misuse scenarios, only for the purpose of taxonomy, risk understanding, and defense analysis. We avoid providing
actionable implementation details beyond what is already available in the cited literature, and we emphasize mitigation, governance, and responsible deployment throughout the paper.

\section*{LLM usage considerations}
This paper studies foundation-model-powered robots, including systems that integrate LLMs, VLMs, and VLAs into robotic perception, planning, policy generation, and embodied execution.
LLMs were used for editorial purposes in this manuscript, and all outputs were inspected by the authors to ensure accuracy and originality.

\ifCLASSOPTIONcompsoc
\else

\fi

\bibliographystyle{IEEEtran}
\bibliography{reference}

\appendix
\setcounter{section}{0}
\renewcommand{\thesection}{\Alph{section}}

\refstepcounter{section}
\section*{\thesection. Coding Attributes}
\label{app:code}


\smallskip
\noindent\textbf{Stage.} 
We consider six stages.
\begin{itemize}[leftmargin=*]
    \item \textbf{Specification}: This stage refers to the early design of the robotic system, including requirement analysis, system specification, architectural design, algorithm selection, and safety-compliance planning.
    \item \textbf{Development}: This stage covers the construction and adaptation of the FM, including dataset construction, model pretraining, fine-tuning, simulation-based training, and adaptation to real-world environments.
    \item \textbf{Integration}: This stage involves assembling and evaluating the FM modules and other components as a complete robotic system. This stage includes functional testing and safety validation across modules.
    \item \textbf{Deployment}: This stage refers to the use of the robotic system in real-world environments, such as task execution and human–robot interaction.
    \item \textbf{Maintenance}: This stage covers post-deployment maintenance, updates and operational management, including model updates, software patching, continual learning, retraining, and performance monitoring.
    \item \textbf{Decommissioning}: This stage concerns the retirement of the robotic system, including data removal, model-weight disposal, hardware recycling, and protection against residual information leakage.
\end{itemize}

\smallskip
\noindent\textbf{System Access.} 
We categorize access into four levels. 
\begin{itemize}[leftmargin=*]
    \item \textbf{None} indicates that system access is not a necessary part of the assumption, such as non-adversarial risks or general privacy exposures.  
    \item \textbf{Black-box} denotes access only through controlled input--output interfaces, such as APIs, queries, or robot interactions.
    \item \textbf{Gray-box} denotes partial visibility into internal signals of a component, such as logits or intermediate activations of FMs, but without full model parameters.
    \item   \textbf{White-box} denotes full access to FMs or system internals, including architecture, parameters, adapters, or training pipelines.
\end{itemize}


\smallskip
\noindent\textbf{Effect.} 
Because the surveyed papers use heterogeneous robots, simulators, metrics, threat models, and privacy settings, we treat these labels as qualitative evidence-strength indicators rather than directly comparable numerical scores. The labels are assigned based only on evidence explicitly reported in the original papers, such as attack success rate, latency, and overhead, where applicable. We code three aspects. 

\textit{Efficacy} measures how strongly a risk, attack, or mitigation achieves its intended objective under the paper's own evaluation setting. 
\begin{itemize}[leftmargin=*]
\item \textbf{High} indicates consistently strong attack success, substantial task degradation, clear privacy leakage, or strong mitigation performance; 
\item \textbf{Medium} indicates clear but limited effectiveness in scope, transferability, or evaluation breadth; 
\item \textbf{Low} indicates weak, preliminary, or only partially demonstrated effects.
\end{itemize}

\textit{Stealth} measures how inconspicuous or difficult to notice an attack or privacy leakage is in the evaluated setting. 
\begin{itemize}[leftmargin=*]
\item \textbf{High} indicates covert, hard-to-detect, or naturally blended behavior; 
\item \textbf{Medium} indicates that stealth is plausible or partially evaluated but not strongly established; \item \textbf{Low} indicates that the attack is visually or operationally obvious, or that stealth is not supported by the paper.
\end{itemize}

\textit{Utility cost} measures the side effect of a mitigation on benign system behavior. 
\begin{itemize}[leftmargin=*]
\item \textbf{High} indicates substantial degradation in clean-task performance, usability, latency, or deployment practicality; 
\item \textbf{Medium} indicates moderate overhead or performance loss; \item \textbf{Low} indicates that the mitigation preserves normal functionality with only minor cost.
\end{itemize}




\refstepcounter{section}

  \begin{figure}[tt]
    \centering
    \includegraphics[width=1\columnwidth]{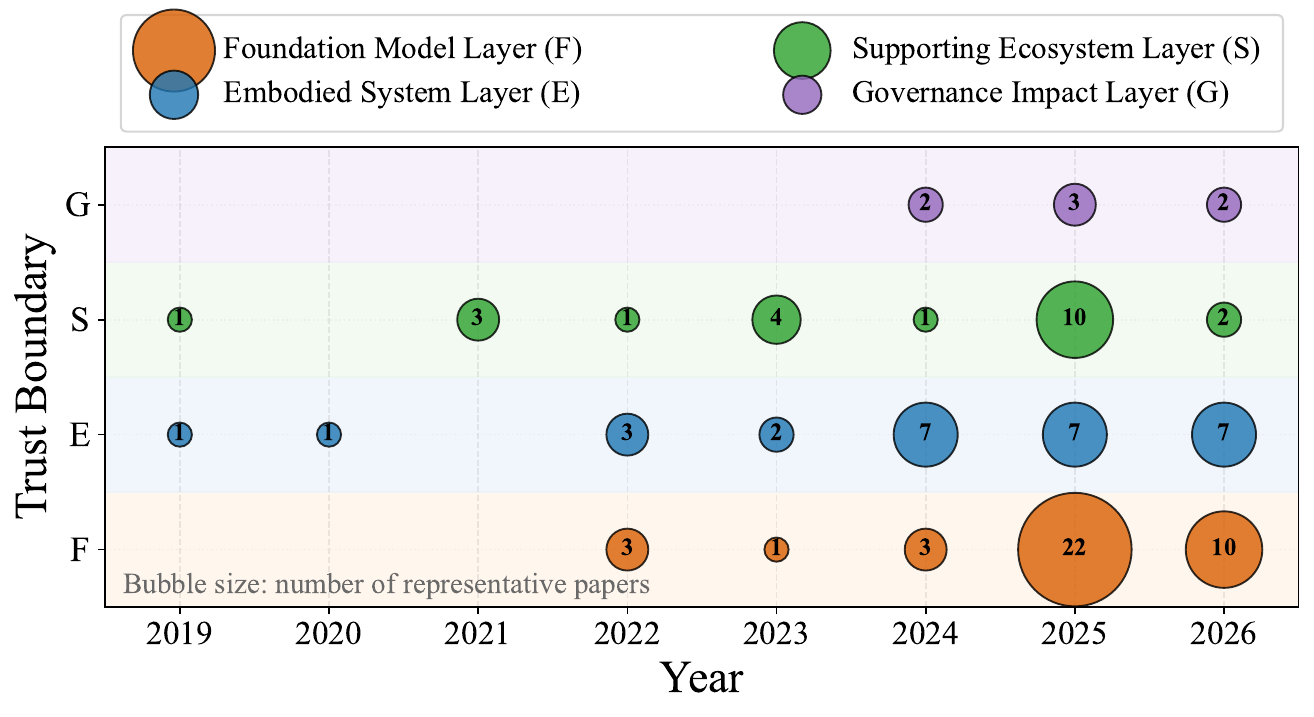}
    \caption{Representative literature timeline of FM-powered robots. The figure groups prior work by four trust boundaries (F/E/S/G). Bubble size denotes the number of representative papers in each year–layer cell. Most studies emerged in the last 3 years, with especially rapid growth at the F and E layers, compared with relatively limited evidence at the ecosystem and governance levels.} 
    \label{fig:time}
\end{figure}

\section*{\thesection. Literature Retrieval and Coding Details}
\label{app:methodology}

To support the systematization in this SoK, we constructed a structured representative corpus through literature retrieval, screening, and coding. Candidate papers were collected from Google Scholar, arXiv, IEEE Xplore, ACM Digital Library, and major security, privacy, robotics, and embodied-AI venues, including IEEE S\&P, USENIX Security, ACM CCS, NDSS, ICRA, IROS, RSS, and CoRL. We further expanded the candidate set through backward and forward citation tracing from relevant surveys, benchmark papers, attack/risk studies, and defense papers.

\noindent\textbf{Retrieval protocol.}
Because terminology in FM-powered robotics is still evolving, we used iterative query expansion rather than a single fixed search string. Consistent with our taxonomy, retrieval terms covered three dimensions: (i) FM-powered robotic systems, e.g., \texttt{LLM}, \texttt{VLM}, \texttt{VLA}, \texttt{foundation model}, \texttt{embodied AI}, \texttt{robot}, and \texttt{robotic manipulation}; (ii) S\&P risks and mitigations, e.g., \texttt{security}, \texttt{privacy}, \texttt{jailbreak}, \texttt{prompt injection}, \texttt{backdoor}, \texttt{poisoning}, \texttt{adversarial attack}, \texttt{membership inference}, \texttt{runtime guardrail}, and \texttt{privacy-preserving}; and (iii) ecosystem and governance domains, e.g., \texttt{cloud robotics}, \texttt{supply chain}, \texttt{telemetry}, \texttt{multi-robot}, \texttt{auditing}, and \texttt{accountability}. Newly identified terms, venues, and citation links were folded back into subsequent search rounds.

After collecting the initial candidates, we removed duplicates and clearly irrelevant works, screened the remaining papers by title and abstract, and checked the full text when necessary. We retained papers that study S\&P for FM-powered robots, as well as robot S\&P studies that are directly transferable to FM-powered robotic systems. Papers with only tangential relevance or insufficient technical detail were excluded. The final coded corpus spans all four trust boundaries and is used for taxonomy construction rather than as an exhaustive bibliometric census. Table~\ref{tab:screening-statistics} summarizes the screening process.

\noindent\textbf{Coding consistency.}
Each retained paper was coded in two passes. The first pass recorded descriptive attributes from the paper's own threat model, target system, method, and evaluation setting. The second pass normalized these attributes into the F-E-S-G codebook and checked consistency across papers with similar mechanisms. Ambiguous cross-layer cases were independently reviewed by at least two authors and resolved through discussion. We assigned the primary layer according to where the risk or mitigation is introduced.

\input{Table/FM_and_Datasets}

\begin{table}[t]

\centering
\caption{Summary of literature retrieval and screening.}
\label{tab:screening-statistics}
\setlength{\tabcolsep}{20pt}
\scriptsize
\begin{tabular}{l c}

\toprule
\textbf{Screening stage} & \textbf{Number of papers} \\
\midrule
 Initial candidates & 290 \\                                                                                                                                                   
  After duplicate/relevance filtering & 214 \\                                                                                                                                  
  After title/abstract screening & 150 \\                                                                                                                                       
  After full-text screening & 118 \\                                                                                                                                            
  Final coded corpus & 96 \\                 
\bottomrule
\end{tabular}

\end{table}

\refstepcounter{section}
\section*{\thesection. Scope and Validity Considerations}
\label{app:scope-validity}
Our systematization has two scope limitations. First, the corpus is structured and representative rather than an exhaustive bibliometric census. It is intended to support taxonomy construction and cross-layer synthesis, not to count every adjacent robotics, AI-safety, or privacy paper. Second, the literature is uneven: Security work is more mature than privacy work, and ecosystem- and governance-level evidence remains comparatively sparse. We therefore treat F-E-S-G as a diagnostic lens for the current field rather than a fixed taxonomy, and expect categories such as VLA safety, ecosystem security, privacy-preserving evidence, and governance accountability to evolve as the literature matures.


\input{Table/E_table}

\input{Table/S_table}
\end{document}

%% file: Table/survey_com.tex
\begin{table*}[t]
  \centering
  \scriptsize
  \caption{Comparison of recent review works on security and privacy in FM-powered robots.}
  \label{tab:fm_robot_survey_compare}

  \resizebox{\textwidth}{!}{%
  \begin{tabular}{@{}l c c *{16}{c}@{}}
  \toprule
  \multirow{3}{*}{\makecell{Related literature}} &
  \multirow{3}{*}{\makecell{FM$^\ddagger$}} &
  \multirow{3}{*}{\makecell{Cross-layer\\synthesis$^\dagger$}} &
  \multicolumn{8}{c}{Security$^\#$} &
  \multicolumn{8}{c}{Privacy$^\#$} \\
  \cmidrule(lr){4-11} \cmidrule(lr){12-19}
  & & &
  \multicolumn{2}{c}{Model (F)} &
  \multicolumn{2}{c}{Embodiment (E)} &
  \multicolumn{2}{c}{Ecosystem (S)} &
  \multicolumn{2}{c}{Governance (G)} &
  \multicolumn{2}{c}{Model (F)} &
  \multicolumn{2}{c}{Embodiment (E)} &
  \multicolumn{2}{c}{Ecosystem (S)} &
  \multicolumn{2}{c}{Governance (G)} \\
  \cmidrule(lr){4-5} \cmidrule(lr){6-7}
  \cmidrule(lr){8-9} \cmidrule(lr){10-11}
  \cmidrule(lr){12-13} \cmidrule(lr){14-15}
  \cmidrule(lr){16-17} \cmidrule(lr){18-19}
  & & &
  \makecell{Atk.} & \makecell{Def.} &
  \makecell{Atk.} & \makecell{Def.} &
  \makecell{Atk.} & \makecell{Def.} &
  \makecell{Atk.} & \makecell{Def.} &
  \makecell{Risk} & \makecell{Def.} &
  \makecell{Risk} & \makecell{Def.} &
  \makecell{Risk} & \makecell{Def.} &
  \makecell{Risk} & \makecell{Def.} \\
  \midrule

  \makecell[l]{Safety at Scale \cite{ma2026safety}} &
  \makecell{L/V} &
  \emptysquare &
  \fullsquare & \fullsquare &
  \emptysquare & \emptysquare &
  \emptysquare & \emptysquare &
  \emptysquare & \emptysquare &
  \fullsquare & \fullsquare &
  \emptysquare & \emptysquare &
  \emptysquare & \emptysquare &
  \emptysquare & \emptysquare \\

  \makecell[l]{Emerged S\&P of LLM Agent \cite{he2025emerged}} &
  L &
  \emptysquare &
  \fullsquare & \halfsquare &
  \emptysquare & \emptysquare &
  \emptysquare & \emptysquare &
  \emptysquare & \emptysquare &
  \halfsquare & \halfsquare &
  \emptysquare & \emptysquare &
  \emptysquare & \emptysquare &
  \emptysquare & \emptysquare \\

  \makecell[l]{Trust in LLM Robotics \cite{huang2025trust}} &
  L &
  \halfsquare &
  \fullsquare & \halfsquare &
  \fullsquare & \fullsquare &
  \emptysquare & \emptysquare &
  \emptysquare & \emptysquare &
  \emptysquare & \emptysquare &
  \emptysquare & \emptysquare &
  \emptysquare & \emptysquare &
  \emptysquare & \emptysquare \\

  \makecell[l]{LLM \& VLM for Robot \cite{hu2025large}} &
  \makecell{L/V} &
  \halfsquare &
  \halfsquare & \halfsquare &
  \halfsquare & \halfsquare &
  \emptysquare & \emptysquare &
  \emptysquare & \emptysquare &
  \halfsquare & \halfsquare &
  \halfsquare & \halfsquare &
  \emptysquare & \emptysquare &
  \emptysquare & \emptysquare \\

  \makecell[l]{Safety of VLA Models \cite{yuan2026safety}} &
  V &
  \halfsquare &
  \fullsquare & \fullsquare &
  \halfsquare & \halfsquare &
  \emptysquare & \emptysquare &
  \emptysquare & \emptysquare &
  \halfsquare & \halfsquare &
  \emptysquare & \emptysquare &
  \emptysquare & \emptysquare &
  \emptysquare & \emptysquare \\

  \makecell[l]{Robust \& Secure Embodied AI \cite{xing2025towards}} &
  \makecell{L/V} &
  \halfsquare &
  \emptysquare & \emptysquare &
  \fullsquare & \fullsquare &
  \fullsquare & \fullsquare &
  \emptysquare & \emptysquare &
  \emptysquare & \emptysquare &
  \fullsquare & \fullsquare &
  \fullsquare & \fullsquare &
  \emptysquare & \emptysquare \\

  \makecell[l]{Safety of Embodied Navigation \cite{wang2025safety}} &
  \makecell{L/V} &
  \emptysquare &
  \emptysquare & \emptysquare &
  \fullsquare & \halfsquare &
  \emptysquare & \emptysquare &
  \emptysquare & \emptysquare &
  \emptysquare & \emptysquare &
  \halfsquare & \halfsquare &
  \emptysquare & \emptysquare &
  \emptysquare & \emptysquare \\

  \makecell[l]{Embodied AI Security \cite{ma2026breaks}} &
  L &
  \halfsquare &
  \emptysquare & \emptysquare &
  \fullsquare & \halfsquare &
  \halfsquare & \halfsquare &
  \emptysquare & \emptysquare &
  \emptysquare & \emptysquare &
  \halfsquare & \halfsquare &
  \halfsquare & \halfsquare &
  \emptysquare & \emptysquare \\

  \makecell[l]{Security Risks in Robotics \cite{efa2024evaluating}} &
  L &
  \halfsquare &
  \emptysquare & \emptysquare &
  \halfsquare & \emptysquare &
  \emptysquare & \emptysquare &
  \halfsquare & \halfsquare &
  \emptysquare & \emptysquare &
  \halfsquare & \emptysquare &
  \emptysquare & \emptysquare &
  \halfsquare & \halfsquare \\

  \makecell[l]{LLMs for Multi-Robot Systems \cite{li2025large}} &
  L &
  \emptysquare &
  \emptysquare & \emptysquare &
  \emptysquare & \emptysquare &
  \emptysquare & \emptysquare &
  \emptysquare & \emptysquare &
  \emptysquare & \emptysquare &
  \halfsquare & \emptysquare &
  \emptysquare & \emptysquare &
  \emptysquare & \emptysquare \\

  \makecell[l]{FM-Driven Robotics \cite{khan2025foundation}} &
  \makecell{L/V} &
  \emptysquare &
  \emptysquare & \emptysquare &
  \emptysquare & \emptysquare &
  \emptysquare & \emptysquare &
  \emptysquare & \emptysquare &
  \emptysquare & \emptysquare &
  \emptysquare & \emptysquare &
  \emptysquare & \emptysquare &
  \emptysquare & \emptysquare \\

  \makecell[l]{Cybersecurity Assessment \cite{surve2025sok}} &
  None &
  \emptysquare &
  \emptysquare & \emptysquare &
  \emptysquare & \emptysquare &
  \fullsquare & \fullsquare &
  \emptysquare & \emptysquare &
  \emptysquare & \emptysquare &
  \emptysquare & \emptysquare &
  \halfsquare & \halfsquare &
  \emptysquare & \emptysquare \\

  \makecell[l]{Secure Robotics \cite{haskard2025secure}} &
  None &
  \halfsquare &
  \emptysquare & \emptysquare &
  \emptysquare & \emptysquare &
  \fullsquare & \fullsquare &
  \halfsquare & \halfsquare &
  \emptysquare & \emptysquare &
  \emptysquare & \emptysquare &
  \halfsquare & \halfsquare &
  \halfsquare & \halfsquare \\

  \makecell[l]{A Survey of Embodied AI \cite{duan2022survey}} &
  None &
  \emptysquare &
  \emptysquare & \emptysquare &
  \emptysquare & \emptysquare &
  \emptysquare & \emptysquare &
  \emptysquare & \emptysquare &
  \emptysquare & \emptysquare &
  \emptysquare & \emptysquare &
  \emptysquare & \emptysquare &
  \emptysquare & \emptysquare \\

  \makecell[l]{Risks for Policy Action \cite{perlo2025embodied}} &
  None &
  \halfsquare &
  \emptysquare & \emptysquare &
  \emptysquare & \emptysquare &
  \emptysquare & \emptysquare &
  \halfsquare & \fullsquare &
  \emptysquare & \emptysquare &
  \emptysquare & \emptysquare &
  \emptysquare & \emptysquare &
  \fullsquare & \fullsquare \\

  \makecell[l]{FMs in Robotics \cite{firoozi2025foundation}} &
  \makecell{L/V} &
  \emptysquare &
  \emptysquare & \emptysquare &
  \emptysquare & \halfsquare &
  \emptysquare & \emptysquare &
  \emptysquare & \emptysquare &
  \emptysquare & \emptysquare &
  \emptysquare & \emptysquare &
  \emptysquare & \emptysquare &
  \emptysquare & \emptysquare \\

  \makecell[l]{Embodied AI \cite{lisondra2026embodied}} &
  \makecell{L/V/A} &
  \halfsquare &
  \emptysquare & \emptysquare &
  \emptysquare & \halfsquare &
  \emptysquare & \halfsquare &
  \emptysquare & \fullsquare &
  \emptysquare & \emptysquare &
  \halfsquare & \halfsquare &
  \halfsquare & \halfsquare &
  \halfsquare & \fullsquare \\

  \textbf{Ours} &
  \makecell{\textbf{L/V/A}} &
  \textbf{\fullsquare} &
  \textbf{\fullsquare} & \textbf{\fullsquare} &
  \textbf{\fullsquare} & \textbf{\fullsquare} &
  \textbf{\fullsquare} & \textbf{\fullsquare} &
  \textbf{\fullsquare} & \textbf{\fullsquare} &
  \textbf{\fullsquare} & \textbf{\fullsquare} &
  \textbf{\fullsquare} & \textbf{\fullsquare} &
  \textbf{\fullsquare} & \textbf{\fullsquare} &
  \textbf{\fullsquare} & \textbf{\fullsquare} \\

  \bottomrule
  \end{tabular}%
  }

  \vspace{1mm}
  \begin{minipage}{1\textwidth}
  \scriptsize
  \raggedright
  \sloppy
  \setlength{\parskip}{0.2pt}
  \setlength{\baselineskip}{8.2pt}

  \noindent\textbf{Cross-layer synthesis$^\dagger$:}
  indicates whether a work analyzes how risks, defenses, or harms propagate,
  amplify, or mismatch across layers.

  \noindent\textbf{FM$^\ddagger$:}
  L = LLM, V = VLM, A = VLA, None = FMs are not a primary organizing focus.

  \noindent\textbf{Security$^\#$:}
  F = foundation model layer; E = embodied system layer;
  S = supporting ecosystem layer; G = governance impact layer.
  Security is coded by whether a work discusses attacks or defenses at each layer, while privacy is coded by whether a work discusses privacy risks or defenses at each layer. 

 \noindent\textbf{Coverage depth:}
\emptysquare = absent or only passing mention;
\halfsquare = mentioned or discussed but not fully systematized;
\fullsquare = systematic framework/taxonomy with substantial coverage.

  \end{minipage}
\end{table*}

%% file: Table/F_table.tex
\begin{table*}[t]
\renewcommand{\arraystretch}{0.9}
\centering
\scriptsize
\caption{Representative Risk and Mitigation Studies Across Security and Privacy Domains at the Foundation-Model Layer (F)}
\label{tab:F}
\setlength{\tabcolsep}{4pt}
\begin{tabular}{c | c | l l l l l l l l l}

\toprule
\multicolumn{1}{c|}{\multirow{3}{*}{\textbf{Domain}}} &
\multicolumn{1}{c|}{\multirow{3}{*}{\textbf{Category}}} &
\multicolumn{1}{c}{\multirow{3}{*}{\textbf{Study}}} &
\multicolumn{1}{c}{\multirow{3}{*}{\textbf{Year}}} &
\multicolumn{1}{c}{\multirow{3}{*}{\textbf{FM}}} &
\multicolumn{1}{c}{\multirow{3}{*}{\textbf{\makecell[c]{Target}}}} &
\multicolumn{1}{c}{\multirow{3}{*}{\textbf{\makecell[c]{Mechanism}}}} &
\multicolumn{1}{c}{\multirow{3}{*}{\textbf{Stage}}} &
\multicolumn{1}{c}{\multirow{3}{*}{\textbf{\makecell[c]{Access}}}} &
\multicolumn{2}{c}{\textbf{Effect}} \\
\cmidrule(lr){10-11}
& & & & & & & & &
\textbf{Efficacy} &
\textbf{\makecell[c]{Stealth/Cost}} \\
\midrule

\multirow{31}{*}{\textbf{Security}} &
\multirow{24}{*}{\textbf{Risks}}& ICL Backdoor Attack \cite{liu2024compromising}
& 2025 & LLM & Planning Module & Model Compromise & Deployment & Black-box & High & High \\

& & State Backdoor \cite{guo2026state}
& 2026 & VLA & Policy Module & Model Compromise & Development & White-box & High & High \\

& & SilentDrift \cite{xu2026silentdrift}
& 2026 & VLA & Policy Module & Model Compromise & Development & White-box & High & High \\

& & GoBA \cite{zhou2025goal}
& 2025 & VLA & Policy Module & Model Compromise & Development & White-box & High & High \\

& & BadVLA \cite{zhou2025badvla}
& 2025 & VLA & Policy Module & Model Compromise & Development & White-box & High & High \\

& & INFUSE \cite{zhou2026inject}
& 2026 & VLA & Policy Module & Model Compromise & Development & White-box & High & High \\

& & BALD \cite{jiao2024can}
& 2025 & LLM & Planning Module & Model Compromise & Development & White-box & High & High \\

& & BEAT \cite{zhan2025visual}
& 2025 & VLM & Planning Module & Model Compromise & Development & White-box & High & High \\

\cmidrule(lr){3-11}

& & RoboPAIR \cite{robey2025jailbreaking}
& 2025 & LLM & Planning Module & Semantic Manipulation & Deployment & Black-box & High & Medium \\

& & BadRobot \cite{zhang2024badrobot}
& 2024 & LLM & Planning Module & Semantic Manipulation & Deployment & Black-box & High & High \\

& & Poex \cite{lu2024poex}
& 2024 & LLM & Planning Module & Semantic Manipulation & Deployment & Black-box & High & Medium \\

& & BadNAVer \cite{lyu2025badnaver}
& 2025 & VLM & Planning Module & Semantic Manipulation & Deployment & Black-box & High & Medium \\

& & Adversarial Attacks \cite{jones2025adversarial}
& 2025 & VLA & Policy Module & Semantic Manipulation & Deployment & White-box & High & Medium \\

& & \makecell[l]{SABER \cite{wu2026saber}}
& 2026
& VLA
& Policy Module
& \makecell[l]{Semantic Manipulation}
& Deployment
& Black-box
& High
& High \\

& & CrossInject \cite{wang2025manipulating}
& 2025 & VLM & Planning Module & Semantic Manipulation & Deployment & Black-box & Medium & Medium \\

& & Prompt Injection \cite{zhang2024study}
& 2024 & LLM & Planning Module & Semantic Manipulation & Deployment & Black-box & Medium & Medium \\

\cmidrule(lr){3-11}

& & Exploring \cite{wang2025exploring}
& 2025 & VLA & Policy Module & Visual Manipulation & Deployment & White-box & High & Medium \\

& & UPA-RFAS \cite{lu2025robots}
& 2025 & VLA & Policy Module & Visual Manipulation & Deployment & Black-box & High & Medium \\

& & PhysPatch \cite{guo2026physpatch}
& 2026 & VLM & Planning Module & Visual Manipulation & Deployment & Black-box & High & Medium \\

& & Tex3D \cite{chen2026tex3d}
& 2026 & VLA & Policy Module & Visual Manipulation & Deployment & White-box & High & High \\

& & TRAP \cite{huang2026trap}
& 2026 & VLA & Planning Module & Visual Manipulation & Deployment & White-box & High & Medium \\

& & CHAI \cite{burbano2025chai}
& 2025 & VLM & Planning Module & Visual Manipulation & Deployment & None & High & Medium \\

& & PI3D \cite{li2026extended}
& 2026 & VLM & Planning Module & Visual Manipulation & Deployment & None & Medium & Medium \\

\cmidrule(lr){3-11}

& & Reward Gaming \cite{skalse2022defining}
& 2022 & - & Policy Module & Misalignment & Specification & None & Medium & High \\

& & Goal Misgeneralization \cite{di2022goal}
& 2022 & - & Policy Module & Misalignment & Development & None & Medium & High \\
\cmidrule(lr){2-11}

& \multirow{7}{*}{\textbf{Mitigation}} & RobustVLA \cite{zhang2025robustvla}
& 2025 & VLA & Policy Module & Model Hardening & Development & White-box & High & Medium \\

& & TREAD \cite{kuramshin2025task}
& 2025 & VLA & Policy Module & Model Hardening & Development & White-box & Medium & Low \\

& & Model-agnostic \cite{xu2025model}
& 2025 & VLA & Perception Module & Model Hardening & Development & White-box & High & Low \\

& & RETAIN \cite{yadav2025robust}
& 2025 & VLA & Policy Module & Model Hardening & Maintenance & White-box & High & Low \\

& & MergeVLA \cite{fu2025mergevla}
& 2025 & VLA & Policy Module & Model Hardening & Maintenance & White-box & High & Medium \\

& & MITD \cite{sahoo2025horcrux}
& 2025 & - & Planning Module & Model Hardening & Development & White-box & Medium & Medium \\

& & SAFE-Dict \cite{wen2026concept}
& 2026 & VLA & Policy Module & Execution Guardrail & Deployment & Gray-box & High & Low \\

\midrule

\multirow{8}{*}{\textbf{Privacy}} & \multirow{5}{*}{\textbf{Risks}} &
VLM-MIA \cite{hu2025membership}
& 2025 & VLM & Planning Module & Membership Inference & Development & Black-box & High & Medium \\

& &  VLA-MIA \cite{peng2026membership}
& 2026 & VLA & Policy Module & Membership Inference & Development & Black-box & High & Medium \\

& &  PRoP \cite{christie2025fine}
& 2025 & - & Policy Module &Personalization Leakage  & Maintenance & Black-box & High & High \\

& & TidyBot \cite{wu2023tidybot}
& 2023 & - & Planning Module & Personalization Leakage & Maintenance & None & Medium & High \\

& & Agentic Surgical AI \cite{zhan2025agentic}
& 2025 & VLA & Policy Module & Fingerprinting Leakage & Development & Gray-box & High & Medium \\

\cmidrule(lr){2-11}

 & \multirow{3}{*}{\textbf{Mitigation}}
& PRoP \cite{christie2025fine}
& 2025 & - & Policy Module & Key-based Gating & Maintenance & White-box & High & Low \\

& & FedVLN \cite{zhou2022fedvln}
& 2022 & VLM & Planning Module & Federated Safeguard & Development & White-box & High & Low \\

& & FedVLA \cite{miao2025fedvla}
& 2025 & VLA & Policy Module & Federated Safeguard & Development & White-box & High & Low \\

\bottomrule
\end{tabular}
\end{table*}

%% file: Table/FM_and_Datasets.tex
\begin{table}[t]
\centering
\caption{Representative FM-powered robotic systems.}
\label{tab:fm-robot-systems}
\scriptsize
\setlength{\tabcolsep}{0.5pt}
\renewcommand{\arraystretch}{1}

\begin{threeparttable}
\begin{tabularx}{\columnwidth}{
llll}
\toprule
\textbf{Name} & \textbf{Planning Module} & \textbf{Policy Module} & \textbf{Avail.} \\
\midrule

\multicolumn{4}{@{}l}{\textit{LLM-based robotic systems}} \\[1pt]
SayCan~\cite{ahn2022can}
& PaLM
& Behavior cloning policy
& Partial \\

Code as Policies~\cite{liang2023code}
& Codex (code-davinci-002) 
& Control primitive APIs
& Partial \\

\makecell[l]{ChatGPT Robotics~\cite{vemprala2024chatgpt}}
& \makecell[l]{ChatGPT (code generation)}
& Pre-defined control APIs
& Partial \\

VoxPoser~\cite{huang2023voxposer}
& \makecell[l]{GPT-4/3.5\\(value-map composition)}
& \makecell[l]{Trajectory optimization\\and MPC}
& Public \\

\midrule
\multicolumn{4}{@{}l}{\textit{VLM-based robotic systems}} \\[1pt]
PaLM-E~\cite{driess2023palm}
& PaLM-E
& RT-1
& Closed \\

SuSIE~\cite{black2024zero}
& InstructPix2Pix
& GCBC policy
& Public \\

PIVOT~\cite{nasiriany2024pivot}
& GPT-4V/Gemini
& Iterative visual optimizer
& Partial \\

ViLa~\cite{hu2023look}
& GPT-4V
& Pre-defined control APIs
& Partial \\

MA~\cite{duan2024manipulate}
& GPT-V/Qwen-VL
& Behavior cloning policy
& Public \\

\midrule
\multicolumn{4}{@{}l}{\textit{VLA-based robotic systems}} \\[1pt]
RT-1~\cite{brohan2022rt}
& SayCan (PaLM)
& RT-1
& Partial \\

RT-2~\cite{zitkovich2023rt}
& \makecell[l]{RT-2 (PaLI-X/PaLM-E)}
& RT-2
& Closed \\

Octo~\cite{team2024octo}
& --
& Octo
& Public \\

OpenVLA~\cite{kim2024openvla}
& \makecell[l]{OpenVLA (Prismatic-7B)}
& OpenVLA
& Public \\

$\pi_0$~\cite{black2024pi}
& $\pi_0$ (PaliGemma)
& $\pi_0$
& Public \\

CogACT~\cite{li2024cogact}
& Llama-2
& DiT
& Public \\

RDT-1B~\cite{liu2024rdt}
& T5-XXL + SigLIP
& RDT-1B
& Public \\

Gemini Robotics~\cite{team2025gemini}
& Gemini Robotics
& Gemini Robotics
& Closed \\

\bottomrule
\end{tabularx}
\begin{tablenotes}[flushleft]
\scriptsize
\item[] \textbf{Availability}: \emph{Public} = reusable code, weights, data, or benchmarks are public; \emph{Partial} = some artifacts are public but key models, APIs, or platforms remain closed; \emph{Closed} = key components are proprietary or vendor-controlled.
\end{tablenotes}
\end{threeparttable}
\end{table}

%% file: Table/E_table.tex
\begin{table*}[t]
\renewcommand{\arraystretch}{0.9}
\centering
\scriptsize
\caption{Representative Risk and Mitigation Studies Across Security and Privacy Domains at the Embodied-System Layer (E)}
\label{tab:E}
\setlength{\tabcolsep}{1pt}
\resizebox{\textwidth}{!}{
\begin{tabular}{c | c | l l l l l l l l l}

\toprule
\multicolumn{1}{c|}{\multirow{3}{*}{\textbf{Domain}}} &
\multicolumn{1}{c|}{\multirow{3}{*}{\textbf{Category}}} &
\multicolumn{1}{c}{\multirow{3}{*}{\textbf{Study}}} &
\multicolumn{1}{c}{\multirow{3}{*}{\textbf{Year}}} &
\multicolumn{1}{c}{\multirow{3}{*}{\textbf{FM}}} &
\multicolumn{1}{c}{\multirow{3}{*}{\textbf{\makecell[c]{Target}}}} &
\multicolumn{1}{c}{\multirow{3}{*}{\textbf{\makecell[c]{Mechanism}}}} &
\multicolumn{1}{c}{\multirow{3}{*}{\textbf{Stage}}} &
\multicolumn{1}{c}{\multirow{3}{*}{\textbf{\makecell[c]{Access}}}} &
\multicolumn{2}{c}{\textbf{Effect}} \\
\cmidrule(lr){10-11}
& & & & & & & & &
\textbf{Efficacy} &
\textbf{\makecell[c]{Stealth/Cost}} \\
\midrule

\multirow{20}{*}{\textbf{Security}} &
\multirow{9}{*}{\textbf{Risks}} &
Blindfold \cite{huang2026jailbreaking}
& 2026 & LLM & Planning Module & Semantic Manipulation & Deployment & Black-box & High & High \\

& & Altered Thoughts \cite{trinh2026altered}
& 2026 & VLA & Policy Module & Semantic Manipulation & Deployment & Gray-box & High & High \\

\cmidrule(lr){3-11}

& & FlyTrap \cite{lu2025flytrap}
& 2025 & - & Perception Module & Visual Manipulation & Deployment & None & Medium & Low \\

& & AdvGrasp \cite{wang2025advgrasp}
& 2025 & - & Execution Module & Visual Manipulation & Deployment & None & High & Medium \\

\cmidrule(lr){3-11}

& & Phantom Menace \cite{lu2026phantom}
& 2026 & VLA & Perception Module & Signal Manipulation & Deployment & None & High & Medium \\

& & Acoustic Manipulation \cite{cheng2023adversarial}
& 2023 & - & Perception Module & Signal Manipulation & Deployment & None & High & Medium \\

& & LiDAR Spoofing \cite{sato2025realism}
& 2025 & - & Perception Module & Signal Manipulation & Deployment & None & High & Medium \\

\cmidrule(lr){3-11}

& & Secure ROS2 \cite{deng2022security}
& 2022 & - & Middleware Module & Middleware Compromise & Deployment & Black-box & High & Medium \\

& & Credential Masquerading \cite{diluoffo2019credential}
& 2019 & - & Middleware Module & Middleware Compromise & Deployment & White-box & High & High \\

\cmidrule(lr){2-11}

& \multirow{11}{*}{\textbf{Mitigation}} &
SAFE \cite{gu2025safe}
& 2025 & VLA & Policy Module & Runtime Guardrail & Deployment & White-box & High & Low \\

& & AHA \cite{duan2024aha}
& 2024 & VLM & Execution Module & Runtime Guardrail & Deployment & Black-box & High & Medium \\

& & Guardian \cite{pacaud2025guardian}
& 2025 & VLM & Execution Module & Runtime Guardrail & Deployment & Black-box & High & Medium \\

& & Vision-Language Models \cite{luo2024vision}
& 2024 & VLM & Execution Module & Runtime Guardrail & Deployment & Black-box & High & Low \\

& & FPC-VLA \cite{yang2025fpc}
& 2025 & VLA & Policy Module & Runtime Guardrail & Deployment & Black-box & High & Medium \\

& & Sentinel \cite{agia2024unpacking}
& 2024 & - & Policy Module & Runtime Guardrail & Deployment & Black-box & High & Low \\

& & LogicGuard \cite{criticslogicguard}
& 2026 & LLM & Planning Module & Runtime Guardrail & Deployment & Black-box & High & Medium \\

& & Safety Guardrails \cite{ravichandran2026safety}
& 2026 & LLM & Planning Module & Runtime Guardrail & Deployment & Black-box & High & Medium \\

& & RoVer \cite{dai2025rover}
& 2025 & VLA & Policy Module & Runtime Guardrail & Deployment & Gray-box & High & Medium \\

& & Ask Before You Act \cite{karli2025ask}
& 2025 & VLA & Policy Module & Runtime Guardrail & Deployment & Gray-box & High & Low \\

& & The Security Analysis \cite{yang2023security}
& 2023 & - & Middleware Module & Middleware Hardening & Integration & White-box & Medium & Low \\

\midrule

\multirow{8}{*}{\textbf{Privacy}} &
\multirow{3}{*}{\textbf{Risks}} &
Robot Vacuum Cleaner \cite{sami2020spying}
& 2020 & - & Perception Module & Eavesdropping & Deployment & None & High & High \\

& & Fingerprinting Robot Movements \cite{shah2022fingerprinting}
& 2022 & - & Kinematics Control Module & Eavesdropping & Deployment & None & High & High \\

& & Formal Analysis \cite{yang2024formal}
& 2024 & - & Middleware Module & Eavesdropping & Deployment & Gray-box & Medium & Medium \\

\cmidrule(lr){2-11}

& \multirow{5}{*}{\textbf{Mitigation}} &
Inherently Privacy-Preserving Vision \cite{taras2024inherently}
& 2024 & - & Perception Module & Data-Minimizing Perception & Specification & None & High & Medium \\

& & Confidant \cite{tang2022confidant}
& 2022 & - & Planning Module & Privacy-Aware Planning & Deployment & None & Medium & Low \\

& & PANav \cite{yu2024panav}
& 2024 & VLM & Planning Module & Privacy-Aware Planning & Deployment & None & Medium & Medium \\

& & FedHIP \cite{cai2024fedhip}
& 2024 & - & Perception Module & Federated Safeguard & Development & None & Medium & Medium \\

& & FTI-SLAM \cite{liu2024fti}
& 2024 & - & Perception Module & Federated Safeguard & Development & None & Medium & Medium \\

\bottomrule
\end{tabular}
}
\end{table*}

%% file: Table/S_table.tex
\begin{table*}[t]
\renewcommand{\arraystretch}{0.9}
\centering
\scriptsize
\caption{Representative Risk and Mitigation Studies Across Security and Privacy Domains at the Supporting-Ecosystem Layer (S)}
\label{tab:S}
\setlength{\tabcolsep}{1pt}
\resizebox{\textwidth}{!}{
\begin{tabular}{c | c | l l l l l l l l l}

\toprule
\multicolumn{1}{c|}{\multirow{3}{*}{\textbf{Domain}}} &
\multicolumn{1}{c|}{\multirow{3}{*}{\textbf{Category}}} &
\multicolumn{1}{c}{\multirow{3}{*}{\textbf{Study}}} &
\multicolumn{1}{c}{\multirow{3}{*}{\textbf{Year}}} &
\multicolumn{1}{c}{\multirow{3}{*}{\textbf{FM}}} &
\multicolumn{1}{c}{\multirow{3}{*}{\textbf{\makecell[c]{Target}}}} &
\multicolumn{1}{c}{\multirow{3}{*}{\textbf{\makecell[c]{Mechanism}}}} &
\multicolumn{1}{c}{\multirow{3}{*}{\textbf{Stage}}} &
\multicolumn{1}{c}{\multirow{3}{*}{\textbf{\makecell[c]{Access}}}} &
\multicolumn{2}{c}{\textbf{Effect}} \\
\cmidrule(lr){10-11}
& & & & & & & & &
\textbf{Efficacy} &
\textbf{\makecell[c]{Stealth/Cost}} \\
\midrule

\multirow{17}{*}{\textbf{Security}} &
\multirow{12}{*}{\textbf{Risks}} &
Cybersecurity AI \cite{mayoral2025}
& 2025 & - & \makecell[l]{External Supporting Infrastructure} & Software-Service Abuse & Deployment & Gray-box & High & High \\

& & Cybersecurity of Humanoid Robot \cite{mayoral2025cybersecurity}
& 2025 & - & \makecell[l]{External Supporting Infrastructure} & Software-Service Abuse & Deployment & Gray-box & High & High \\

& & Scanning for ROS \cite{demarinis2019scanning}
& 2019 & - & Middleware Module & Software-Service Abuse & Deployment & Black-box & Medium & Low \\

\cmidrule(lr){3-11}

& & TrojanRobot \cite{wang2025trojanrobot}
& 2025 & VLM & Perception Module & Supply-Chain Compromise & Development & White-box & High & High \\

& & Supply Chain \cite{sakib2025supply}
& 2025 & - & Middleware Module & Supply-Chain Compromise & Development & White-box & High & High \\

& & From Prompt to Physical Action \cite{xie2026prompt}
& 2026 & LLM & Planning Module & Supply-Chain Compromise & Development & White-box & High & High \\

\cmidrule(lr){3-11}

& & From Prompts to Motors \cite{shaikh2025prompts}
& 2025 & LLM & \makecell[l]{External Supporting Infrastructure} & Man-in-the-Middle Attack & Deployment & Gray-box & High & Medium \\

& & Net-GPT \cite{piggott2023net}
& 2023 & LLM & \makecell[l]{External Supporting Infrastructure} & Man-in-the-Middle Attack & Deployment & Black-box & High & High \\

\cmidrule(lr){3-11}

& & InfectBot \cite{huang2026propagating}
& 2026 & LLM & \makecell[l]{External Supporting Infrastructure} & Multi-Robot System Attack & Deployment & Black-box & High & High \\

& & Byzantine Threats \cite{deng2021investigation}
& 2021 & - & \makecell[l]{External Supporting Infrastructure} & Multi-Robot System Attack & Deployment & Gray-box & High & Medium \\

& & DoS Attacks \cite{xu2021novel}
& 2021 & - & \makecell[l]{External Supporting Infrastructure} & Multi-Robot System Attack & Deployment & Gray-box & High & Medium \\

& & Raven \cite{yeke2025automated}
& 2025 & - & Policy Module & Multi-Robot System Attack & Deployment & Gray-box & High & High \\

\cmidrule(lr){2-11}

& \multirow{5}{*}{\textbf{Mitigation}} &
From Prompt to Physical Action \cite{xie2026prompt}
& 2026 & LLM & Planning Module & Supply-Chain Verification & Deployment & Black-box & Medium & High \\

& & RoboRebound \cite{gandhi2025roborebound}
& 2025 & - & Execution Module & Multi-Robot Resilience & Deployment & White-box & High & Medium \\

& & Byzantine \cite{wardega2023byzantine}
& 2023 & - & \makecell[l]{External Supporting Infrastructure} & Multi-Robot Resilience & Deployment & None & High & Low \\

& & Byzantine Robots \cite{strobel2023robot}
& 2023 & - & \makecell[l]{External Supporting Infrastructure} & Multi-Robot Resilience & Deployment & None & High & Low \\

& & Trusted Operations \cite{santoso2023trusted}
& 2023 & - & \makecell[l]{External Supporting Infrastructure} & Runtime MITM Detection & Deployment & Gray-box & High & Low \\

\midrule

\multirow{7}{*}{\textbf{Privacy}} &
\multirow{4}{*}{\textbf{Risks}} &
Investigating the Privacy \cite{ulsmaag2024investigating}
& 2024 & - & \makecell[l]{External Supporting Infrastructure} & Traffic-Analysis Leakage & Deployment & None & Medium & High \\

& & Privacy-Preserving Robotic Perception \cite{antonazzi2025privacy}
& 2025 & - & \makecell[l]{External Supporting Infrastructure} & Data-Outsourcing Leakage & Deployment & None & High & High \\

& & Can You Still See Me? \cite{shah2022can}
& 2022 & - & \makecell[l]{External Supporting Infrastructure} & Traffic-Analysis Leakage & Deployment & None & High & High \\

& & On the Feasibility \cite{tang2025feasibility}
& 2025 & - & \makecell[l]{External Supporting Infrastructure} & Traffic-Analysis Leakage & Deployment & None & High & High \\

\cmidrule(lr){2-11}

& \multirow{3}{*}{\textbf{Mitigation}} &
Privacy-Preserving Robotic Perception \cite{antonazzi2025privacy}
& 2025 & - & \makecell[l]{External Supporting Infrastructure} & Data-Outsourcing Protection & Deployment & None & High & Medium \\

& & Privacy-Preserving Face Recognition \cite{karri2021privacy}
& 2021 & - & \makecell[l]{External Supporting Infrastructure} & Data-Outsourcing Protection & Deployment & None & Medium & Medium \\

& & RoboComm \cite{singh2025robocomm}
& 2025 & - & \makecell[l]{External Supporting Infrastructure} & Inter-Robot Privacy Protection & Deployment & None & Medium & Low \\

\bottomrule
\end{tabular}
}
\end{table*}